\documentclass{article}

\PassOptionsToPackage{numbers, compress}{natbib}
\usepackage[final]{neurips_2024}

\usepackage[utf8]{inputenc} 
\usepackage[T1]{fontenc}    
\usepackage{hyperref}       
\usepackage{url}            
\usepackage{booktabs}       
\usepackage{amsfonts}       
\usepackage{nicefrac}       
\usepackage{microtype}      

\usepackage[table,dvipsnames]{xcolor}
\usepackage{makecell}

\usepackage{algorithm}
\usepackage[noend]{algorithmic}

\usepackage{amsmath}
\usepackage{amssymb}
\usepackage{mathtools}
\usepackage{amsthm}
\usepackage{accents}
\newcommand\unbar[1]{%
	\underaccent{\bar}{#1}}
\usepackage[capitalize,noabbrev]{cleveref}
\theoremstyle{plain}
\newtheorem{theorem}{Theorem}[section]
\newtheorem{proposition}[theorem]{Proposition}
\newtheorem{lemma}[theorem]{Lemma}
\newtheorem{corollary}[theorem]{Corollary}
\theoremstyle{definition}
\newtheorem{definition}[theorem]{Definition}
\newtheorem{assumption}[theorem]{Assumption}
\theoremstyle{remark}

\title{Unveiling the Potential of Robustness in Selecting Conditional Average Treatment Effect Estimators}

\author{%
  Yiyan Huang \thanks{The co-first authors.}\\
  The Hong Kong Polytechnic University \\
  \texttt{yiyhuang3-c@my.cityu.edu.hk} \\
   \And
   Cheuk Hang Leung \footnotemark[1] \\
   City University of Hong Kong \\
   \texttt{chleung87@cityu.edu.hk} \\
   \AND
   Siyi Wang \\
   City University of Hong Kong \\
   \texttt{siyi.wang@my.cityu.edu.hk} \\
   \And
   Yijun Li \\
   City University of Hong Kong \\
   \texttt{yijunli5-c@my.cityu.edu.hk} \\
   \And
   Qi Wu \thanks{Corresponding author.} \\
   City University of Hong Kong \\
   \texttt{qi.wu@cityu.edu.hk} \\
}

\begin{document}

\maketitle

\begin{abstract}
The growing demand for personalized decision-making has led to a surge of interest in estimating the Conditional Average Treatment Effect (CATE). Various types of CATE estimators have been developed with advancements in machine learning and causal inference. However, selecting the desirable CATE estimator through a conventional model validation procedure remains impractical due to the absence of counterfactual outcomes in observational data. Existing approaches for CATE estimator selection, such as plug-in and pseudo-outcome metrics, face two challenges. First, they must determine the metric form and the underlying machine learning models for fitting nuisance parameters (e.g., outcome function, propensity function, and plug-in learner). Second, they lack a specific focus on selecting a robust CATE estimator. To address these challenges, this paper introduces a Distributionally Robust Metric (DRM) for CATE estimator selection. The proposed DRM is nuisance-free, eliminating the need to fit models for nuisance parameters, and it effectively prioritizes the selection of a distributionally robust CATE estimator. The experimental results validate the effectiveness of the DRM method in selecting CATE estimators that are robust to the distribution shift incurred by covariate shift and hidden confounders.
\end{abstract}
\section{Introduction}\label{sec:intro}
The escalating demand for decision-making has sparked an increasing interest in \textit{Causal Inference} across various research domains, such as economics \citep{farrell2015robust, chernozhukov2018double, kitagawa2018should, abadie2023should}, statistics \citep{wager2018estimation, li2022random, foster2023orthogonal, kennedy2023towards}, healthcare \citep{zhang2019medical, foster2011subgroup, qian2021synctwin, bica2021real, kinyanjui2022adcb}, and financial application \citep{bottou2013counterfactual, chu2021graph, huang2021causal, donnelly2021correction, huang2023towards, fernandez2023comparison, li2024causal}. The primary goal in personalized decision-making is to quantify the causal effect of a specific treatment (or policy/intervention) on the target outcome, and understanding such causal effects is closely connected with identifying the \textit{Conditional Average Treatment Effect (CATE)}. In observational studies, identifying the CATE faces a significant and fundamental challenge: the absence of \textit{counterfactual} knowledge. According to Rubin Causal Model \citep{rubin2005causal}, the CATE is determined by comparing \textit{potential outcomes} under different treatment assignments (i.e., treat and control) for a specific covariate. Nonetheless, in real-world applications, we can only observe the potential outcome under the actual treatment (i.e., \textit{factual outcome}), while the potential outcome under the alternative treatment (i.e., \textit{counterfactual outcome}) remains unobserved. The unavailability of the counterfactual outcome is widely recognized as the fundamental problem in causal inference \citep{holland1986statistics}, making it difficult to accurately determine the true value of the CATE.


The advancement of machine learning (ML) has opened up a promising opportunity to improve the CATE estimation from observational data. Several innovative CATE estimation approaches, such as meta-learners and causal ML models, have been proposed to tackle the fundamental challenge in causal inference and enhance the predictive accuracy of CATE estimates (as discussed in Section \ref{sec:related}). Nevertheless, the emergence of various CATE estimation methods has brought forth a new question: \textbf{\textit{Given multifarious options for CATE estimators, which should be chosen?}} In observational data, treatment is often non-random and propensity scores remain unknown. Conventional model validation procedures, unfortunately, are not suitable for CATE estimator selection in this case due to the absence of ground truth CATE labels. Therefore, exploring proper metrics for CATE estimator selection remains an essential yet challenging research topic in causal inference.

Recent research has emphasized the significance of model selection for CATE estimators, as highlighted in \cite{schuler2018comparison, pmlr-v202-curth23b, mahajan2022empirical}. These works have proposed and summarized two types of criteria for CATE estimator selection, the \textit{plug-in} metric $\mathcal{R}_{\tilde{\tau}}^{plug}(\hat{\tau})$ and the \textit{pseudo-outcome} metric $\mathcal{R}_{\tilde{Y}}^{pseudo}(\hat{\tau})$:
\begin{equation}
	\begin{aligned}
		\mathcal{R}_{\tilde{\tau}}^{plug}(\hat{\tau})=\sqrt{\frac{1}{n} \sum_{i=1}^{n} (\hat{\tau}(X_i) - \tilde{\tau}(X_i))^2}, \quad 
		\mathcal{R}_{\tilde{Y}}^{pseudo}(\hat{\tau})=\sqrt{\frac{1}{n} \sum_{i=1}^{n} (\hat{\tau}(X_i) - \tilde{Y}_i)^2}. \label{eqt:plug_pseudo}
	\end{aligned}
\end{equation}
One can establish a plug-in estimator $\tilde{\tau}$ or construct a pseudo-outcome estimator $\tilde{Y}$ using the validation data to select CATE estimator $\hat{\tau}$. The previous studies \cite{schuler2018comparison, pmlr-v202-curth23b, mahajan2022empirical} have shown that these metrics offer some assistance in identifying well-performing CATE estimators. However, two additional challenges are still encountered in these two metrics.

\paragraph{\textit{Challenge 1: How to determine the metric form and underlying ML models for nuisance parameters?}} As previously discussed, plug-in and pseudo-outcome metrics have various forms, and both of them rely on estimating nuisance parameters $\tilde{\eta}$ using ML algorithms such as linear models, tree-based models, etc. Plug-in metrics even need to fit an additional ML model for the plug-in learner $\tilde{\tau}$. However, selecting the suitable metric form and ML algorithms can be very difficult without the knowledge of true data generating process. Consequently, we might go round in circles as this challenge leads us back to the original estimator selection problem \citep{pmlr-v202-curth23b}.

\paragraph{\textit{Challenge 2: These metrics are not well-targeted for selecting robust a CATE estimator.}} In potential outcome framework \citep{rubin2005causal}, the factual distribution $P^{F}$ and the counterfactual distribution $P^{CF}$ for $t \in \{0,1\}$ can be defined as follows: 
\begin{equation}
	\begin{aligned}
		P^{F}:=P(X,Y^t|T=t)&=P(Y^t|X,T=t)P(X|T=t);\\
		P^{CF}:=P(X,Y^t|T=1-t)&=P(Y^t|X,T=1-t)P(X|T=1-t).
	\end{aligned}\label{eqt:PF_PCF}
\end{equation}
The above \eqref{eqt:PF_PCF} reveals that the covariate shift $P(X|T=t) \neq P(X|T=1-t)$ leads to a distribution shift between $P^{F}$ and $P^{CF}$ - and such distribution shift can be further exacerbated once the unconfoundedness assumption $P(Y^t|X,T=t)=P(Y^t|X,T=1-t)$ is violated. It is widely recognized that ML models often struggle when the training and test data do not adhere to the same distribution. Therefore, it becomes essential to select a CATE estimator learned on $P^{F}$ that demonstrates robust performance to the counterfactual distribution $P^{CF}$. This need for robustness holds even greater significance than the pursuit of an ideal ``stellar'' estimator because striving for the perfect estimator can be futile in the absence of ground truth counterfactual labels.




\paragraph{Contributions.} In this paper, we propose a Distributionally Robust Metric (DRM) for CATE estimator selection. The main contributions are summarized as follows: (1) The proposed DRM method is nuisance-free, eliminating the need to fit models for nuisance parameters (outcome function, propensity function, and plug-in learner). (2) The DRM method is designed to prioritize selecting a distributionally robust CATE estimator. (3) We provide a finite sample analysis of the proposed distributionally robust value $\hat{\mathcal{V}}^t(\hat{\tau})$ for $t \in \{0, 1\}$, showing it decays to $\mathcal{V}^t(\hat{\tau})$ at a rate of $n^{-1/2}$. (4) Experimental results validate the effectiveness of the DRM method in selecting a CATE estimator that is robust to the distribution shift incurred by covariate shift and hidden confounders.

\section{Background of CATE Estimator Selection} \label{sec:background}
Suppose the observational data contain $n$ i.i.d. samples $\{(x_i, t_i, y_i)\}_{i=1}^{n}$, with the associated random variables being $\{(X_i, T_i, Y_i)\}_{i=1}^{n}$. For each unit $i$, $X_i \in \mathcal{X} \subset \mathbb{R}^d$ is $d$-dimensional covariates and $T_i \in \{0,1\}$ is the binary treatment. Potential outcomes for treat ($T=1$) and control ($T=0$) are denoted by $Y^1, Y^0 \in \mathcal{Y} \subset \mathbb{R}$. The observed (factual) outcome is $Y=TY^1 + (1-T)Y^0$. The propensity score \citep{rosenbaum1983central} is defined as $\pi(x):=P(T=1 \mid X=x)$. The conditional mean potential outcome surface is defined as $\mu_t(x) :=\mathbb{E}\left[Y^t \mid X=x\right]$ for $t \in \{0,1\}$. The true CATE is defined as
\begin{align*}
	\tau_{true}(x) :=\mathbb{E}\left[Y^1-Y^0 \mid X=x\right]=\mu_1(x)-\mu_0(x).
\end{align*}

Following the standard and necessary assumptions in potential outcome framework \citep{rubin2005causal}, we impose Assumption \ref{assump} that ensure treatment effects are identifiable.
\begin{assumption}[Consistency, Overlap, and Unconfoundedness] \label{assump} Consistency: If the treatment is $t$, then the observed outcome $Y$ equals $Y^t$. Overlap: The propensity score is bounded away from $0$ to $1$, i.e., $0<\pi(x)<1$, $\forall x \in \mathcal{X}$. Unconfoundedness \footnote{Note that in the setting C of our experiments, the unconfoundedness assumption is violated, leading to misspecified nuisance parameters in CATE estimators and baseline selectors.}: $Y^t \perp\!\!\!\perp T \mid X, \; \forall t \in \{0, 1\}$.
\end{assumption}

The goal of CATE estimator selection is to select the best CATE estimator, denoted by $\hat{\tau}_{best}$, from a set of $J$ candidate estimators $\{\hat{\tau}_1, \dots, \hat{\tau}_J\}$:
\begin{align}
	\hat{\tau}_{best}=\mathop{\arg \min}_{\hat{\tau} \in \{\hat{\tau}_1, \dots, \hat{\tau}_J\}} \mathcal{R}^{oracle}(\hat{\tau}), \quad \mathcal{R}^{oracle}(\hat{\tau}):=\sqrt{\frac{1}{n}\sum_{i=1}^{n}(\hat{\tau}(X_i)-\tau_{true}(X_i))^2}. \label{eqt:pehe}
\end{align}
Here, $\mathcal{R}^{oracle}(\hat{\tau})$ is associated with $\mathbb{E}[(\hat{\tau}(X)-\tau_{true}(X))^2]$, known as the Precision of Estimating Heterogeneous Effects (PEHE) w.r.t. $\hat{\tau}$ \citep{hill2011bayesian, shalit2017estimating}. Note that $\mathcal{R}^{oracle}(\hat{\tau})$ cannot be employed to evaluate CATE estimators' performances in real applications as we do not have access to $\tau_{true}$. Previous studies have introduced plug-in and pseudo-outcome metrics to aid in CATE estimator selection, as shown in equation \eqref{eqt:plug_pseudo}. Then, the CATE estimator $\hat{\tau}_{select}$ is selected on validation data by
\begin{equation}
	\hat{\tau}_{select}= \mathop{\arg \min}_{\hat{\tau} \in \{\hat{\tau}_1, \dots, \hat{\tau}_J\}} \mathcal{R}_{\tilde{\tau}}^{plug}(\hat{\tau}) \quad \text{or} \quad \hat{\tau}_{select}=\mathop{\arg \min}_{\hat{\tau} \in \{\hat{\tau}_1, \dots, \hat{\tau}_J\}}\mathcal{R}_{\tilde{Y}}^{pseudo}(\hat{\tau}).
\end{equation} 
Notably, both the plug-in and pseudo-outcome metrics necessitate the fitting of nuisance parameters $\tilde{\eta}$ (e.g., $\tilde{\eta}=(\tilde{\mu}_1, \tilde{\mu}_0, \tilde{\pi})$) using off-the-shelf ML models. While some papers like \cite{cui2024selective} address the selection of nuisance parameters for Averate Treatment Effect (ATE) estimators, e.g., the doubly robust estimator \cite{farrell2015robust, chernozhukov2018double, huang2022robust}, our paper focuses on the selection of CATE estimators rather than nuisance parameters. For the plug-in metric, $\tilde{\tau}$ can be constructed using any CATE estimator discussed in Appendix \ref{sec:CATE Learners}, yielding metrics such as plug-T, plug-DR, etc. For the pseudo-outcome metric, $\tilde{Y}$ can be constructed using a specific formula discussed in Appendix \ref{sec:CATE selectors}, yielding metrics such as pseudo-DR, pseudo-R, etc. The metrics based on the influence function \citep{alaa2019validating} and the R-learner objective \citep{nie2021quasi} are categorized into the pseudo-outcome metric. The categorization of plug-in and pseudo-outcome metrics maintains consistency with \cite{pmlr-v202-curth23b, mahajan2022empirical}. 

\section{Related Work} \label{sec:related}
\paragraph{CATE estimation.}
Recent advancements in ML have emerged as powerful tools for estimating CATE from observational data, and researchers pay particular attention to \textit{meta-learners} and \textit{causal ML} models. 
Existing meta-learners mainly include traditional learners such as S-learner, T-learner, PS-learner, and IPW-learner, as well as new learners such as X-learner \citep{kunzel2019metalearners}, U-learner \citep{fisherinverse, nie2021quasi}, DR-learner \citep{kennedy2023towards, foster2023orthogonal}, R-learner \citep{nie2021quasi}, and RA-learner \citep{curth2021nonparametric}. The specific details of these meta-learners are stated in Appendix \ref{sec:CATE Learners}. Additionally, some studies also focus on developing innovative causal ML models for CATE estimation, such as Causal BART \citep{hahn2020bayesian}, Causal Forest \citep{wager2018estimation, athey2019generalized, oprescu2019orthogonal}, generative models like CEVAE \citep{louizos2017causal} and GANITE \citep{yoon2018ganite}, representation learning nets including SITE \citep{yao2018representation}, TARNet \citep{shalit2017estimating}, Dragonnet \citep{shi2019adapting}, FlexTENet \citep{curth2021inductive}, and HTCE \citep{bica2022transfer}, disentangled learning nets like D$^2$VD \citep{kuang2017treatment, kuang2020data}, DeR-CFR \citep{wu2022learning}, and DR-CFR \citep{hassanpour2019learning}, and representation balancing nets such as BNN \citep{johansson2016learning}, CFRNet \citep{shalit2017estimating}, DKLITE \citep{zhang2020learning}, IGNITE \citep{guo2021ignite}, BWCFR \citep{assaad2021counterfactual}, DRRB \citep{huang2023towards}, and DIGNet \citep{huang2024dignet}. Recent surveys \citep{guo2020survey, yao2021survey, nogueira2022methods} have also conducted a systematic review of various causal inference methods.

\paragraph{CATE estimator selection.}
Compared to the diverse range of CATE estimation methods, selecting CATE estimators has received limited attention in existing causal inference research.
Current methods for selecting CATE estimators can be broadly classified into two main categories.
\textbf{The first category}, which is also considered in this paper, involves using plug-in and pseudo-outcome methods to evaluate CATE estimators. These methods share two common characteristics: 1) Both methods require fitting ML models for nuisances (e.g., outcome function, propensity function, CATE function) on a validation set and then implementing the learned ML models in either the plug-in surrogate or the pseudo-outcome surrogate; 2) Both methods serve as surrogates for the expected error between the CATE estimator and the true CATE, i.e., $\mathcal{R}^{oracle}(\hat{\tau})$ in equation \eqref{eqt:pehe}. The difference between the two methods is that the plug-in method directly approximates the true CATE function, where only covariate variables are involved, while the pseudo-outcome method typically constructs a specific formula incorporating covariates, treatment, and outcome variables. For example, the pseudo-DR proposed in \cite{saito2020counterfactual} is constructed by the outcome predictors learned with representation balancing objective \cite{shalit2017estimating, johansson2022generalization}. Recent research \citep{schuler2018comparison, pmlr-v202-curth23b, mahajan2022empirical} has conducted thorough empirical investigations into exploring these two methods for selecting CATE estimators. Their findings suggest that no single selection criterion can universally outperform others in all scenarios in the task of selecting CATE estimators. More details of the two selection methods are stated in Appendix \ref{sec:CATE selectors}. 
\textbf{The second category} considers leveraging the data generating process (DGP) to generate synthetic data with the known true CATE function, allowing the validation of CATE estimators' performance on this synthetic data. For example, authors in \cite{advani2019mostly} find that placebo and structured empirical Monte Carlo methods are helpful for estimator selection under some restrictive conditions. In addition, researchers in \cite{schuler2017synth, athey2021using, parikh2022validating} focus on training generative models to enforce the generated data to approximate the distribution of the observed data. However, the DGP-based method still faces some limitations in CATE estimator selection: 1) it only guarantees the resemblance of the generated data to the factual distribution, without considering the counterfactual distribution; and ii) there is a potential risk of the method favoring estimators that closely resemble the generative models \citep{curth2021really}.

\section{The Distributionally Robust Metric} \label{sec:DRE_theorem}
In this section, we introduce the Distributionally Robust Metric (DRM) for CATE estimator selection. First, we capture the uncertainty in PEHE in a distributionally robust manner (Section \ref{sec:pehe upper bound}). We then establish the DRM based on the distributionally robust value of PEHE (Section \ref{sec:estbalish DRM}).
\subsection{Capturing the Uncertainty in PEHE} \label{sec:pehe upper bound}
\begin{proposition}\label{prop:pehe_decompose} The PEHE w.r.t. the CATE estimator $\hat{\tau}$ can be decomposed as follows:
	\begin{equation}
		\begin{aligned}
			\mathbb{E}[(\hat{\tau}(X)-\tau_{true}(X))^2]
			= \mathbb{E}[\hat{\tau}(X)^2] + 2 \mathbb{E}[\hat{\tau}(X)Y^0] + 2 \mathbb{E}[-\hat{\tau}(X)Y^1] + \zeta,
		\end{aligned}\label{eqt:pehe_decompose}
	\end{equation}
	where $\zeta = \mathbb{E}[(\mu_1(X)-\mu_0(X))^2]$.
	The proof is deferred to Appendix \ref{app:proof_pehe_decompose}.
\end{proposition}
Proposition \ref{prop:pehe_decompose} indicates that the PEHE is equal to four terms, where $\mathbb{E}[\hat{\tau}(X)^2]$,  $\mathbb{E}[\hat{\tau}(X)Y^0]$, and $\mathbb{E}[-\hat{\tau}(X)Y^1]$ depend on $\hat{\tau}$, while $\zeta$ is a constant that is independent of $\hat{\tau}$. The term $\mathbb{E}[\hat{\tau}(X)Y^t]$ for $t \in \{0,1\}$ can be further decomposed as follows:
\begin{equation}
	\begin{aligned}
		\mathbb{E}[\hat{\tau}(X)Y^t]
		=\underbrace{\mathbb{E}[\hat{\tau}(X)Y^t|T=t]}_{\text{(a)  Empirically computable}}P(T=t)
		+ \underbrace{\mathbb{E}[\hat{\tau}(X)Y^t|T=1-t]}_{\text{(b)  Empirically uncomputable}}P(T=1-t).
	\end{aligned}
	\label{eqt:decompose}
\end{equation}

Equation (\ref{eqt:decompose}a) can be computed empirically since the potential outcome $Y^t$ is observable in the group of $T=t$. However, equation (\ref{eqt:decompose}b) is empirically uncomputable due to the unavailability of $Y^t$ in the group of $T=1-t$. The unknown term $\mathbb{E}[\hat{\tau}(X)Y^t|T=1-t]$ therefore determines the uncertainty in PEHE. To capture such an uncertainty, we therefore establish distributionally robust values for $\mathbb{E}[\hat{\tau}(X)Y^0|T=1]$ and $\mathbb{E}[-\hat{\tau}(X)Y^1|T=0]$ based on a Kullback-Leibler (KL) ambiguity set.

\begin{definition}[KL ambiguity set] \label{def:kl_ball} 	
	Given two distributions $Q$ and $P$ and the ambiguity radius $\epsilon > 0$. The KL ambiguity (uncertainty) set $\mathcal{B}_\epsilon(P)$ is defined as
	\begin{equation}
		\begin{aligned}
			\mathcal{B}_\epsilon(P):=\{Q:D_{KL}(Q||P)\leq \epsilon\}, \quad \text{where} \ D_{KL}(Q||P)=\int_{\mathcal{X}}q(x) \log \frac{q(x)}{p(x)}dx.
		\end{aligned}
	\end{equation}
	Here, $D_{KL}(Q||P)$ denotes the KL divergence of some arbitrary distribution $Q$ from the reference distribution $P$. Now we define the distribution of $(X,Y^0,Y^1)$ in the treated and controlled groups as
	\begin{equation}
		\begin{aligned}
			P_{T}:=P(X,Y^0, Y^1|T=1); \ P_{C}:=P(X,Y^0,Y^1|T=0). \label{eqt:PTPC}
		\end{aligned}
	\end{equation}
\end{definition}
By setting an adequately large ambiguity radius in Definition \ref{def:kl_ball}, the following inequalities hold for $\mathbb{E}[\hat{\tau}(X)Y^0|T=1]=\mathbb{E}^{P_T}[\hat{\tau}(X)Y^0]$ and $\mathbb{E}[-\hat{\tau}(X)Y^1|T=0]=\mathbb{E}^{P_C}[-\hat{\tau}(X)Y^1]$:
\begin{equation}
	\begin{aligned}
		\mathbb{E}[\hat{\tau}(X)Y^0|T=1]=\mathbb{E}^{P_T}[\hat{\tau}(X)Y^0] &\leq \sup_{Q \in B_{\epsilon_0}(P_{C})}\mathbb{E}^{Q}[\hat{\tau}(X)Y^0]=: \mathcal{V}^0(\hat{\tau});\\
		\mathbb{E}[-\hat{\tau}(X)Y^1|T=0]=\mathbb{E}^{P_C}[-\hat{\tau}(X)Y^1] &\leq \sup_{Q \in B_{\epsilon_1}(P_{T})} \mathbb{E}^{Q}[-\hat{\tau}(X)Y^1]=:\mathcal{V}^1(\hat{\tau}).
	\end{aligned} \label{eqt:V_def}
\end{equation}
To provide a clearer understanding, let us consider the example of $\mathbb{E}^{P_T}[\hat{\tau}(X)Y^0]$. Since the term $\mathbb{E}[\hat{\tau}(X)Y^0]$ is computable on its factual distribution $P_C$ but uncomputable on its counterfactual distribution $P_T$, we can construct an ambiguity set centered around the distribution $P_C$ such that it is large enough to contain the distribution $P_{T}$. By doing so, we can capture the uncertainty of $\mathbb{E}^{P_T}[\hat{\tau}(X)Y^0]$ w.r.t. $\hat{\tau}$. In other words, the value of the uncomputable quantity $\mathbb{E}^{P_T}[\hat{\tau}(X)Y^0]$ will be \textbf{at most} $\mathcal{V}^0(\hat{\tau})$. Similarly, the value of the uncomputable quantity $\mathbb{E}^{P_C}[-\hat{\tau}(X)Y^1]$ will be \textbf{at most} $\mathcal{V}^1(\hat{\tau})$. Obviously, the uncertainty in PEHE will be larger if the distribution shift between factual and counterfactual distribution is severer. Consequently, we can obtain the distributionally robust value of PEHE in Corollary \ref{cor:upper bound}, which measures the uncertainty in PEHE.
\vspace{0.1cm}
\begin{corollary} \label{cor:upper bound} Let $\mathcal{V}^0(\hat{\tau})$ and $\mathcal{V}^1(\hat{\tau})$ be the quantities defined in equation \eqref{eqt:V_def}, $\zeta$ be the constant given in Proposition \ref{prop:pehe_decompose}, $u_1:=P(T=1)$, and $u_0=1-u_1=P(T=0)$. The distributionally robust value of PEHE w.r.t. $\hat{\tau}$ is defined as $\mathcal{V}_{PEHE}(\hat{\tau})$ such that
	\begin{equation}
		\begin{aligned}
			&\mathbb{E}[(\hat{\tau}(X)-\tau_{true}(X))^2] \leq \mathcal{V}_{PEHE}(\hat{\tau}) \\
			&= \mathbb{E}[\hat{\tau}(X)^2] + 2\left(u_0\mathbb{E}^{P_C}[\hat{\tau}(X)Y^0] + u_1\mathbb{E}^{P_T}[-\hat{\tau}(X)Y^1]\right) 
			+ 2\left(u_0\mathcal{V}^1(\hat{\tau})+u_1\mathcal{V}^0(\hat{\tau})\right) + \zeta.
		\end{aligned}\label{eqt:pehe_upperbound_final}
	\end{equation}
\end{corollary}
\subsection{Establishing Distributionally Robust Metric} \label{sec:estbalish DRM}
As Corollary \ref{cor:upper bound} provides the distributionally robust (worst-case) value of PEHE, it can naturally measure the robustness of the CATE estimator $\hat{\tau}$ against distribution shift between counterfactual distribution and factual distribution. In this section, we will provide two steps involved in using Corollary \ref{cor:upper bound} to construct the DRM method for CATE estimator selection.

\paragraph{Step 1: Establishing computational tractability of $\mathcal{V}^t(\hat{\tau})$.}
The distributionally robust values $\mathcal{V}^0(\hat{\tau})$ and $\mathcal{V}^1(\hat{\tau})$ in equation \eqref{eqt:pehe_upperbound_final} are initially defined as supremum problems over infinite support, presenting a substantial computational challenge. Theorem \ref{thm:dual_form} reformulates the infeasible supremum problems into tractable minimum problems.
\begin{theorem} \label{thm:dual_form} The distributionally robust values $\mathcal{V}^0(\hat{\tau})$ and $\mathcal{V}^1(\hat{\tau})$ in equation \eqref{eqt:V_def} are equivalent to
	\begin{equation}
		\begin{aligned}
			&\mathcal{V}^0(\hat{\tau})=\min_{\lambda_0 > 0} \lambda_0 \epsilon_0 + \lambda_0\log \mathbb{E}^{P_C}[\exp (\hat{\tau}(X)Y^0/\lambda_0)];\label{eqt:dual} \\
			&\mathcal{V}^1(\hat{\tau})=\min_{\lambda_1 > 0} \lambda_1 \epsilon_1 + \lambda_1\log \mathbb{E}^{P_T}[\exp (-\hat{\tau}(X)Y^1/\lambda_1)].
		\end{aligned}
	\end{equation}
	The proof is deferred to Appendix \ref{app:proof_dual}.
\end{theorem}
In the finite-sample scenario, $\mathcal{V}^0(\hat{\tau})$ and $\mathcal{V}^1(\hat{\tau})$ can be empirically approximated as follows:
\begin{equation}
	\begin{aligned}
		&\hat{\mathcal{V}}^0(\hat{\tau})=\min_{\lambda_0 > 0} \lambda_0 \epsilon_0 + \lambda_0\log \frac{1}{n_c}\sum_{i=1}^{n}(1-T_i) \exp(\hat{\tau}(X_i)Y_i/\lambda_0); \\ \label{eqt:V_hat}
		&\hat{\mathcal{V}}^1(\hat{\tau})=
		\min_{\lambda_1 > 0} \lambda_1 \epsilon_1 + \lambda_1\log \frac{1}{n_t}\sum_{i=1}^{n}T_i \exp(-\hat{\tau}(X_i)Y_i/\lambda_1).
	\end{aligned}
\end{equation}
Note that in equation \eqref{eqt:V_hat}, the potential outcomes $Y^0$ and $Y^1$ are replaced by the observed outcome $Y$ due to the fact that $(1-T)Y^0=(1-T)Y$ and $TY^1=TY$, which aligns with the Consistency assumption in Assumption \ref{assump}. We then provide a finite-sample analysis of the gap between $\hat{\mathcal{V}}^t(\hat{\tau})$ and $\mathcal{V}^t(\hat{\tau})$ in the following Theorem \ref{thm:convergence}, which suggests the gap decays at a rate of $n^{-1/2}$.

\begin{theorem}\label{thm:convergence} Let $u_t:=P(T=t)$ for $t \in \{0,1\}$. Assume $0<\unbar{\lambda} \leq \lambda_0, \lambda_1 \leq \bar{\lambda}$ and $\hat{\tau}(X)Y$ is bounded within the range of $\unbar{M}$ to $\bar{M}$. Define $C_{exp}=\mathbf{1}_{\{\unbar{M} \leq \bar{M} \leq 0\}}\exp\left(\bar{M}/\bar{\lambda} - \unbar{M}/\unbar{\lambda} \right)+\mathbf{1}_{\{\unbar{M} \leq 0, \bar{M} \geq 0\}}\exp\left(\bar{M}/\unbar{\lambda} - \unbar{M}/\unbar{\lambda} \right)+\mathbf{1}_{\{0 \leq \unbar{M} \leq \bar{M}\}}\exp\left(\bar{M}/\unbar{\lambda} - \unbar{M}/\bar{\lambda}\right)$. For $n \geq 2/u^2 \log (2/\delta)$ and $t \in \{0,1\}$, with probability $1-\delta$, we have
	\begin{equation}
		\begin{aligned}			
			|\hat{\mathcal{V}}^t(\hat{\tau})-\mathcal{V}^t(\hat{\tau})|
			\leq \mathcal{O}\left(\sqrt{\frac{8 \bar{\lambda}^2 \log\frac{2}{\delta}}{nu_t^2} C_{exp}^2 }\right) + \mathcal{O}\left(\sqrt{\frac{2\bar{\lambda}^2\log(\frac{2}{\delta})}{nu_t^2}} \right).
		\end{aligned}
	\end{equation}
	The proof is deferred to Appendix \ref{app:proof_convergence}.
\end{theorem}

\paragraph{Step 2: Finalizing Distributionally Robust Metric for CATE estimator selection.}
We first define two functions that are useful in obtaining $\mathcal{V}^0(\hat{\tau})$ and $\mathcal{V}^1(\hat{\tau})$:
\begin{subequations}
	\begin{align}
		\begin{split}
			&\hat{F}_0(\lambda_0, \epsilon_0;\hat{\tau})=\lambda_0 \epsilon_0 + \lambda_0\log \frac{1}{n_c}\sum_{i=1}^{n_c} e^{\frac{Z_i}{\lambda_0}}, \ \hat{F}_1(\lambda_1, \epsilon_1;\hat{\tau})=\lambda_1 \epsilon_1 + \lambda_1\log \frac{1}{n_t}\sum_{i=1}^{n_t} e^{\frac{-Z_i}{\lambda_1}}; 
		\end{split}
		\label{eqt:F}
		\\
		\begin{split}
			&\frac{\partial \hat{F}_0}{\partial \lambda_0} = \epsilon_0 + \log \sum_{i=1}^{n_c} \frac{e^{\frac{Z_i}{\lambda_0}}}{n_c}
			-\frac{\sum_{i=1}^{n_c}Z_ie^{\frac{Z_i}{\lambda_0}}}{\lambda_0\sum_{i=1}^{n_c}e^{\frac{Z_i}{\lambda_0}}}, \  \frac{\partial \hat{F}_1}{\partial \lambda_1} = \epsilon_1 + \log \sum_{i=1}^{n_t} \frac{e^{\frac{-Z_i}{\lambda_1}}}{n_t}
			-\frac{\sum_{i=1}^{n_t}-Z_ie^{\frac{-Z_i}{\lambda_1}}}{\lambda_1\sum_{i=1}^{n_t}e^{\frac{-Z_i}{\lambda_1}}}.
		\end{split}
		\label{eqt:F'}
	\end{align}
\end{subequations}\noindent
Here, $Z$ denotes $\hat{\tau}(X)Y$ for notational simplicity. We then use the Newton-Raphson method to find the empirical solution for $\mathcal{\hat{V}}^t(\hat{\tau})$, exploiting the convexity of $\hat{F}_t(\lambda_t, \epsilon_t;\hat{\tau})$ w.r.t. $\lambda_t$. Based on the distributionally robust value of PEHE, i.e., $\mathcal{V}_{PEHE}(\hat{\tau})$ in equation \eqref{eqt:pehe_upperbound_final}, we finally obtain the selected estimator $\hat{\tau}_{select}= \mathop{\arg \min}_{\hat{\tau} \in \{\hat{\tau}_1, \dots, \hat{\tau}_J\}}\mathcal{R}^{DRM}(\hat{\tau})$ such that 
\begin{equation}
	\begin{aligned}
		\mathcal{R}^{DRM}(\hat{\tau}) = \frac{1}{n}\sum_{i=1}^{n} \hat{\tau}(X_i)^2
		+ \frac{2}{n}\left(\sum_{i=1}^{n_c}\hat{\tau}(X_i)Y_i + \sum_{i=1}^{n_t}-\hat{\tau}(X_i)Y_i
		+ n_c\mathcal{\hat{V}}^1(\hat{\tau}) + n_t\mathcal{\hat{V}}^0(\hat{\tau}) \right)   . \label{eqt:DRM}
	\end{aligned}
\end{equation}
Algorithm \ref{alg:DRM} provides complete procedure of using the DRM method for CATE estimator selection.
\begin{algorithm}[t]
	\begin{algorithmic}[1]
		\caption{Using DRM for CATE Estimator Selection}
		\label{alg:DRM}
		\REQUIRE The candidate CATE estimators $\{\hat{\tau}_1, \dots, \hat{\tau}_J\}$. The validation dataset with $n$ i.i.d. observational samples $\{(X_i,T_i,Y_i)\}_{i=1}^{n}$. The number of iterations $K$. The initialization $\lambda_0^{(0)}$ and $\lambda_1^{(0)}$. The ambiguity radius $\epsilon_0$ and $\epsilon_1$.
		\FOR{$j=1$ to $J$}
		\FOR{$k=0$ to $K-1$}
		\STATE Compute $\hat{F}_t(\lambda_t^{(k)}, \epsilon_t;\hat{\tau}_j)$ for $t \in \{0,1\}$ by equation \eqref{eqt:F}.
		\STATE Compute $\partial \hat{F}_t(\lambda_t^{(k)}, \epsilon_t;\hat{\tau}_j) / \partial \lambda_t^{(k)}$ for $t \in \{0,1\}$ by equation \eqref{eqt:F'}.
		\STATE $\lambda_t^{(k+1)}\leftarrow \max \{\lambda_t^{(k)}-\hat{F}_t(\lambda_t^{(k)}, \epsilon_t;\hat{\tau}_j) / (\partial \hat{F}_t(\lambda_t^{(k)}, \epsilon_t;\hat{\tau}_j) / \partial \lambda_t^{(k)}), 0\}$ for $t \in \{0,1\}$.
		\STATE Save $\hat{\mathcal{V}}^t(\hat{\tau}_j)[k]=\hat{F}_t(\lambda_t^{(k+1)}, \epsilon_t;\hat{\tau}_j)$ for $t \in \{0,1\}$.
		\ENDFOR
		\STATE Return $\hat{\mathcal{V}}^t(\hat{\tau}_j)=\arg \min_{ k \in \{0,\dots,K-1\}} \hat{\mathcal{V}}^t(\hat{\tau}_j)[k]$ for $t \in \{0,1\}$.
		\STATE Use $\hat{\mathcal{V}}^0(\hat{\tau}_j)$ and $\hat{\mathcal{V}}^1(\hat{\tau}_j)$ to compute $\mathcal{R}^{DRM}(\hat{\tau}_j)$ by equation \eqref{eqt:DRM}.
		\ENDFOR
		\ENSURE $\hat{\tau}_{select}= \mathop{\arg \min}_{\hat{\tau} \in \{\hat{\tau}_1, \dots, \hat{\tau}_J\}}\mathcal{R}^{DRM}(\hat{\tau})$.
	\end{algorithmic}
\end{algorithm}

\paragraph{Discussion on the ambiguity radius $\epsilon$.}
The ambiguity radius $\epsilon$ plays a critical role in real-world applications \citep{mohajerin2018data, ma2020understanding, pflug2023multistage}. However, determining an appropriate value for $\epsilon$ can be challenging as it requires striking a balance between ensuring the bound in equation \eqref{eqt:V_def} holds and maintaining its tightness. Specifically, if $\epsilon$ is set too small, it fails to guarantee that the counterfactual distribution is contained within the ambiguity set centered at factual distribution (the bound in Corollary \ref{cor:upper bound} can hold). On the other hand, if $\epsilon$ is set too large, even though the ambiguity set can encompass more distributions to ensure the counterfactual distribution is contained, the bound in Corollary \ref{cor:upper bound} can be less tight. In general, selecting a proper ambiguity radius is an open problem in distributioanlly robust optimization (DRO) literature \citep{hu2013kullback, mohajerin2018data, kuhn2019wasserstein, levy2020large, wang2021sinkhorn}.

In this paper, we provide a guidance for determining the ambiguity radius for our DRM method. Based on the above discussion, an ideal radius should be $\epsilon_1^*=D_{KL}(P_C||P_T)$ and $\epsilon_0^*=D_{KL}(P_T||P_C)$, which ensures that the bound in Corollary \ref{cor:upper bound} holds and is tight. However, as defined in equation \eqref{eqt:PTPC}, both $P_C$ and $P_T$ involve counterfactual information, making it unattainable to directly compute $D_{KL}(P_C||P_T)$ and $D_{KL}(P_T||P_C)$. To overcome this challenge, we demonstrate that Proposition \ref{prop:kl_equal} provides an intriguing alternative approach to acquire $D_{KL}(P_C||P_T)$ and $D_{KL}(P_T||P_C)$ when unconfoundedness in Assumption \ref{assump} is satisfied.
\begin{proposition} \label{prop:kl_equal} Let $P^T_X:=P(X|T=1)$ and $P^C_X:=P(X|T=0)$ denote the covariates distribution in the treat and control group, respectively. Assuming that random variables $(X, T, Y^1, Y^0)$ satisfy the unconfoundedness in Assumption \ref{assump}, we have
	\begin{align}
		D_{KL}(P_C||P_T)
		=D_{KL}(P^C_X||P^T_X); \quad
		D_{KL}(P_T||P_C)
		=D_{KL}(P^T_X||P^C_X).
	\end{align}
	The proof is deferred to Appendix \ref{app:proof_kl_equal}.
\end{proposition}
Proposition \ref{prop:kl_equal} provides an important insight that the uncomputable term $D_{KL}(P_C||P_T)$ (or $D_{KL}(P_T||P_C)$) can be replaced by a computable quantity $D_{KL}(P^C_X||P^T_X)$ (or $D_{KL}(P^T_X||P^C_X)$), where $P^C_X$ and $P^T_X$ are empirically observable. Consequently, the ideal ambiguity radius can be set as $\epsilon_1^*=D_{KL}(P^C_X||P^T_X)$ and $\epsilon_0^*=D_{KL}(P^T_X||P^C_X)$. While the KL divergence can be approximated using empirical algorithm (e.g, Nearest-Neighbor \citep{wang2006nearest, noh2014bias}), we recommend setting the ambiguity radius larger than the empirically approximated KL divergence (see specific explanations in Appendix \ref{app:additional_explanation}). This is necessary because it ensures that the ambiguity set is large enough to contain the target distribution. It is also important to note that though the Algorithm \ref{alg:DRM} involves approximating $\epsilon_1^*=D_{KL}(P^C_X||P^T_X)$ and $\epsilon_0^*=D_{KL}(P^T_X||P^C_X)$, the DRM itself remains free of nuisances, as this approach only determines the ambiguity radius but does not involve learning any nuisance function such as the outcome function, propensity function, and plug-in learner.

\section{Experiments}

\subsection{Experimental Setup.}
\paragraph{Estimators \& Selectors.} We consider a total of \textbf{36 CATE estimators}, comprising the combination of 4 base ML models and 9 meta-learners. Specifically, the base ML models are Linear Regression (LR), Support Vector Machine (SVM), Random Forests (RF), and Neural Net (Net). We consider these ML models for CATE estimators because they are representative of both rigid and flexible models, with each encoded distinct inductive biases, as highlighted by \cite{curth2021inductive, pmlr-v202-curth23b}. Note that for the LR method, we employ Ridge regression for regression tasks and Logistic regression for classification tasks. As for the remaining methods, we utilize their corresponding regressors and classifiers for regression and classification tasks, respectively. Regarding the meta-learners, we select a set of both traditional basic learners (S-, T-, PS-, and IPW-learners) and recently developed learners (X-, DR-, U-, R-, and RA-learners), as detailed in Appendix \ref{sec:CATE Learners}.
We consider \textbf{14 CATE selectors}, consisting of 9 plug-in methods that rely on the above 9 learners, 3 pseudo-outcome methods (pseudo-DR, -R, and -IF), the random selection, the factual selection (from the 6-learner pool with S-, T-), the Nearest-Neighbor Matching \cite{rolling2014model}, and our proposed DRM. The specific details of baseline selectors are stated in Appendix \ref{sec:CATE selectors}. We employ the eXtreme Gradient Boosting (XGB) \citep{chen2016xgboost} as the underlying ML model for both plug-in and pseudo-outcome methods. We choose XGB because: i) it demonstrates superior performance in various scenarios, ensuring a good performance of baseline selectors; ii) the need to avoid potential congeniality bias that may arise from using the similar ML models employed in CATE estimators \citep{pmlr-v202-curth23b}; iii) aligning with \cite{alaa2019validating} where XGB is used for their proposed pseudo-IF metric. The details of hyperparameters for nuisance models are stated in Section \ref{app:hyper} of Appendix.


\paragraph{Dataset.} Since the ground truth of CATE is unavailable in real-world data, previous studies commonly utilize semi-synthetic datasets to compare model performance. In line with \cite{curth2021inductive, pmlr-v202-curth23b}, we collect the covariates with $n=4802$ data points from ACIC2016 dataset \citep{dorie2019automated}. Then, we generate treatment with $T_i|X_i \sim Bern(1/(1+\exp(-\xi (\beta_T' X_i+3))))$, where $Bern$ indicates the Bernoulli distribution. The potential outcomes are generated by a linear function with interaction terms:
\begin{equation*}
	\begin{aligned}
		&Y_i=\sum_{j}^{d}\beta'_{j} X_{i;j} + \sum_{j=1}^{d}\sum_{k=j}^{d}\beta'_{j,k} X_{i;j}X_{i;k}+\sum_{j=1}^{d}\sum_{k=j}^{d}\sum_{l=k}^{d}\beta'_{j,k,l} X_{i;j}X_{i;k}X_{i;l} + T_i\sum_{j=1}^{d}\gamma_{j}X_{i;j} + \epsilon_i.
	\end{aligned}
\end{equation*}
The coefficient values are set as follows: $\beta_T, \beta_{j}, \beta_{j,k}, \beta_{j,k,l} \sim Bern(0.2)$, $\gamma_j \sim Bern(\rho)$, and $\epsilon_i \sim \mathcal{N}(0, 0.1)$. The parameter $\xi$ in treatment assignment represents the level of selection bias, and the parameter $\rho$ in $\gamma_j$ represents the complexity of the CATE function. We adopt the above data generating process to randomly generate 100 datasets, each with a training/validation/testing ratio of 49\%/21\%/30\%.

\paragraph{Settings.} In this section, we mainly investigate whether the estimator selected by DRM can demonstrate robustness to the selection bias and unobserved confounders. In addition, as demonstrated in \cite{curth2021inductive, pmlr-v202-curth23b}, the complexity of CATE function also affects relative performance of estimators and selectors. Given these considerations, we design the following three settings to compare the CATE selectors. \textbf{Setting A:} With the unconfoundedness assumption, let $\rho$ vary in $\{0, 0.1, 0.3\}$ with fixing $\xi=1$. \textbf{Setting B:} With the unconfoundedness assumption, let $\xi$ vary in $\{0, 1, 2\}$ with fixing $\rho=0.1$. \textbf{Setting C}: Without unconfoundedness assumption, fix $\rho=0.1$ and $\xi=1$. Then randomly remove $\lfloor m \cdot d \rfloor$ covariates such that the dimension of observed covariates is $d - \lfloor m \cdot d \rfloor$, where $m$ denotes the ratio of missing covariates varying in $\{0.1, 0.5, 0.9\}$.  All the experiments are run on Dell 3640 with Intel Xeon W-1290P 3.60GHz CPU.

\paragraph{Comparison criteria.} The CATE estimator $\hat{\tau}$ is believed better if it achieves a smaller difference between $\mathcal{R}^{oracle}(\hat{\tau})$ and $\mathcal{R}^{oracle}(\hat{\tau}_{best})$, where $\hat{\tau}_{best}$ is the actual best estimator in equation \eqref{eqt:pehe}. We therefore use the following Regret criteria to compare estimators chosen by different selectors:
\begin{equation*}
	\begin{aligned}
		\text{Regret}=\mathcal{R}^{oracle}(\hat{\tau}_{select})-\mathcal{R}^{oracle}(\hat{\tau}_{best}).
	\end{aligned}
\end{equation*}
To further assess the ranking ability of each selector, we calculate the Spearman rank correlation between the rank order determined by the oracle metric $\mathcal{R}^{oracle}(\hat{\tau})$ and the rank order determined by each selector. All the reported values (Mean $\pm$ Standard deviation) are computed over 100 runs.

\subsection{Experimental Results} \label{sec:exp_results}
\begin{table}[t]
	\centering
	\caption{Comparison of Regret for different selectors across Settings A, B, and C (Note that B ($\xi=1$) matches A ($\rho=0.1$)). Reported values (mean $\pm$ standard deviation) are computed over 100 experiments. Bold denotes the best three results among all selectors. Smaller value is better.}
	\resizebox{1\columnwidth}{!}{
		\begin{tabular}{ccccccccc}
			\toprule
			& A ($\rho=0$) & A ($\rho=0.1$) & A ($\rho=0.3$) & B ($\xi=0$) & B ($\xi=2$) & C ($m=0.1$) & C ($m=0.5$) & C ($m=0.9$) \\
			\midrule
			Plug-U & 47.87\small{$\pm$94.89} & 39.22\small{$\pm$60.78} & 32.13\small{$\pm$61.49} & 0.51\small{$\pm$2.09} & 151.98\small{$\pm$291.20} & 39.41\small{$\pm$52.58} & 54.47\small{$\pm$209.86} & 15.58\small{$\pm$26.41} \\
			Plug-S & 3.38\small{$\pm$7.73} & 2.65\small{$\pm$5.65} & 2.25\small{$\pm$5.64} & \textbf{0.22}\small{$\pm$1.08} & \textbf{5.91}\small{$\pm$10.61} & 2.72\small{$\pm$4.70} & 3.74\small{$\pm$6.34} & 4.65\small{$\pm$6.33} \\
			Plug-PS & 3.08\small{$\pm$7.27} & 2.65\small{$\pm$5.65} & 2.24\small{$\pm$5.64} & \textbf{0.22}\small{$\pm$1.08} & \textbf{5.91}\small{$\pm$10.61} & 2.72\small{$\pm$4.70} & 3.55\small{$\pm$6.11} & 4.65\small{$\pm$6.33} \\
			Plug-T & 59.12\small{$\pm$21.87} & 56.38\small{$\pm$23.02} & 55.35\small{$\pm$21.36} & 10.48\small{$\pm$10.72} & 64.96\small{$\pm$18.03} & 59.87\small{$\pm$18.73} & 43.28\small{$\pm$23.64} & 36.28\small{$\pm$19.33} \\
			Plug-X & 8.10\small{$\pm$10.57} & 7.67\small{$\pm$12.26} & 5.77\small{$\pm$11.26} & 4.76\small{$\pm$10.81} & 11.55\small{$\pm$13.94} & 7.78\small{$\pm$15.02} & 10.76\small{$\pm$14.62} & 11.80\small{$\pm$10.82} \\
			Plug-IPW & 33.37\small{$\pm$28.34} & 35.78\small{$\pm$27.50} & 35.26\small{$\pm$27.24} & 4.43\small{$\pm$7.40} & 58.64\small{$\pm$23.85} & 38.43\small{$\pm$31.16} & 24.98\small{$\pm$22.79} & 20.54\small{$\pm$19.67} \\
			Plug-DR & 43.11\small{$\pm$26.54} & 43.76\small{$\pm$26.92} & 44.20\small{$\pm$26.48} & 4.22\small{$\pm$7.97} & 64.60\small{$\pm$18.88} & 46.57\small{$\pm$32.93} & 28.66\small{$\pm$23.08} & 23.59\small{$\pm$18.21} \\
			Plug-R & \textbf{1.92}\small{$\pm$4.91} & \textbf{2.62}\small{$\pm$15.63} & \textbf{1.47}\small{$\pm$3.60} & 0.43\small{$\pm$2.09} & 9.78\small{$\pm$31.43} & \textbf{1.91}\small{$\pm$4.94} & \textbf{2.53}\small{$\pm$6.51} & \textbf{2.12}\small{$\pm$4.94} \\
			Plug-RA & 56.60\small{$\pm$24.06} & 57.69\small{$\pm$19.98} & 54.60\small{$\pm$22.86} & 6.60\small{$\pm$9.16} & 64.50\small{$\pm$17.72} & 55.87\small{$\pm$19.63} & 40.48\small{$\pm$24.55} & 33.34\small{$\pm$19.04} \\
			Pseudo-DR & 61.35\small{$\pm$22.41} & 61.09\small{$\pm$20.08} & 59.06\small{$\pm$19.46} & 14.75\small{$\pm$22.95} & 70.02\small{$\pm$17.53} & 62.08\small{$\pm$19.65} & 48.83\small{$\pm$25.78} & 44.99\small{$\pm$23.53} \\
			Pseudo-R & 9.85\small{$\pm$27.04} & 14.12\small{$\pm$45.74} & 5.94\small{$\pm$21.05} & 4.73\small{$\pm$20.40} & 14.86\small{$\pm$30.64} & 10.58\small{$\pm$24.63} & 15.93\small{$\pm$29.84} & 21.26\small{$\pm$32.51} \\
			Pseudo-IF & 64.54\small{$\pm$15.18} & 62.49\small{$\pm$16.61} & 62.69\small{$\pm$16.13} & 26.73\small{$\pm$23.60} & 65.74\small{$\pm$16.68} & 60.06\small{$\pm$21.16} & 54.96\small{$\pm$20.63} & 38.60\small{$\pm$22.21} \\
			Random & 7214\small{$\pm$22745} & 6511\small{$\pm$21651} & 4196\small{$\pm$17049} & 1135\small{$\pm$5596} & 7549\small{$\pm$22498} & 3768\small{$\pm$16625} & 6214\small{$\pm$19942} & 3445\small{$\pm$14591} \\
			Fact  & 51.09\small{$\pm$18.00} & 50.86\small{$\pm$19.40} & 51.01\small{$\pm$21.03} & 14.33\small{$\pm$16.54} & 65.23\small{$\pm$27.53} & 48.92\small{$\pm$17.19} & 47.40\small{$\pm$22.51} & 40.37\small{$\pm$23.14} \\
			Matching & 60.85\small{$\pm$21.45} & 62.18\small{$\pm$17.77} & 59.91\small{$\pm$18.67} & 13.33\small{$\pm$22.86} & 68.98\small{$\pm$17.24} & 61.52\small{$\pm$19.04} & 52.83\small{$\pm$23.85} & 40.01\small{$\pm$24.05} \\
			DRM   & \textbf{0.96}\small{$\pm$3.67} & \textbf{0.84}\small{$\pm$4.83} & \textbf{1.25}\small{$\pm$5.97} & 0.38\small{$\pm$1.39} & 15.51\small{$\pm$112.62} & \textbf{1.56}\small{$\pm$8.73} & \textbf{1.40}\small{$\pm$8.67} & \textbf{1.26}\small{$\pm$3.52} \\
			\bottomrule
		\end{tabular}%
	}
	\label{tab:reg_all}%
\end{table}%

		\begin{table}[t]
			\centering
			\caption{Comparison of rank correlation for different selectors across Settings A, B, and C (Note that B ($\xi=1$) matches A ($\rho=0.1$)). Bold denotes the best three results among all selectors. Reported values (mean $\pm$ standard deviation) are computed over 100 experiments. Larger is better.}
			\resizebox{1\columnwidth}{!}{
				\begin{tabular}{ccccccccc}
					\toprule
					& A ($\rho=0$) & A ($\rho=0.1$) & A ($\rho=0.3$) & B ($\xi=0$) & B ($\xi=2$) & C ($m=0.1$) & C ($m=0.5$) & C ($m=0.9$) \\
					\midrule
					Plug-U & 0.69\small{$\pm$0.34} & 0.70\small{$\pm$0.35} & 0.75\small{$\pm$0.29} & \textbf{0.95}\small{$\pm$0.04} & 0.53\small{$\pm$0.30} & 0.68\small{$\pm$0.33} & 0.73\small{$\pm$0.34} & 0.83\small{$\pm$0.24} \\
					Plug-S & 0.95\small{$\pm$0.06} & 0.95\small{$\pm$0.06} & 0.95\small{$\pm$0.05} & \textbf{0.95}\small{$\pm$0.04} & \textbf{0.95}\small{$\pm$0.05} & 0.95\small{$\pm$0.03} & 0.95\small{$\pm$0.05} & 0.91\small{$\pm$0.07} \\
					Plug-PS & 0.95\small{$\pm$0.06} & 0.95\small{$\pm$0.06} & 0.95\small{$\pm$0.05} & \textbf{0.95}\small{$\pm$0.04} & \textbf{0.95}\small{$\pm$0.05} & 0.95\small{$\pm$0.03} & 0.95\small{$\pm$0.05} & 0.91\small{$\pm$0.07} \\
					Plug-T & 0.54\small{$\pm$0.18} & 0.54\small{$\pm$0.18} & 0.54\small{$\pm$0.16} & 0.89\small{$\pm$0.07} & 0.57\small{$\pm$0.16} & 0.51\small{$\pm$0.16} & 0.58\small{$\pm$0.21} & 0.59\small{$\pm$0.21} \\
					Plug-X & 0.94\small{$\pm$0.05} & 0.94\small{$\pm$0.04} & 0.94\small{$\pm$0.04} & 0.93\small{$\pm$0.05} & 0.93\small{$\pm$0.05} & 0.93\small{$\pm$0.04} & 0.92\small{$\pm$0.06} & 0.85\small{$\pm$0.13} \\
					Plug-IPW & 0.72\small{$\pm$0.19} & 0.71\small{$\pm$0.19} & 0.71\small{$\pm$0.19} & 0.92\small{$\pm$0.06} & 0.68\small{$\pm$0.15} & 0.69\small{$\pm$0.19} & 0.76\small{$\pm$0.18} & 0.77\small{$\pm$0.17} \\
					Plug-DR & 0.65\small{$\pm$0.19} & 0.63\small{$\pm$0.20} & 0.63\small{$\pm$0.18} & 0.93\small{$\pm$0.06} & 0.59\small{$\pm$0.16} & 0.61\small{$\pm$0.18} & 0.71\small{$\pm$0.21} & 0.73\small{$\pm$0.18} \\
					Plug-R & \textbf{0.96}\small{$\pm$0.03} & \textbf{0.96}\small{$\pm$0.03} & \textbf{0.96}\small{$\pm$0.03} & \textbf{0.95}\small{$\pm$0.04} & 0.93\small{$\pm$0.07} & \textbf{0.96}\small{$\pm$0.03} & \textbf{0.96}\small{$\pm$0.05} & \textbf{0.96}\small{$\pm$0.04} \\
					Plug-RA & 0.55\small{$\pm$0.19} & 0.54\small{$\pm$0.17} & 0.55\small{$\pm$0.17} & 0.92\small{$\pm$0.06} & 0.57\small{$\pm$0.15} & 0.53\small{$\pm$0.17} & 0.60\small{$\pm$0.22} & 0.62\small{$\pm$0.21} \\
					Pseudo-DR & 0.54\small{$\pm$0.18} & 0.53\small{$\pm$0.18} & 0.53\small{$\pm$0.16} & 0.87\small{$\pm$0.10} & 0.55\small{$\pm$0.15} & 0.50\small{$\pm$0.17} & 0.54\small{$\pm$0.24} & 0.58\small{$\pm$0.23} \\
					Pseudo-R & 0.86\small{$\pm$0.11} & 0.87\small{$\pm$0.09} & 0.88\small{$\pm$0.08} & 0.93\small{$\pm$0.06} & 0.83\small{$\pm$0.13} & 0.85\small{$\pm$0.13} & 0.85\small{$\pm$0.12} & 0.80\small{$\pm$0.16} \\
					Pseudo-IF & 0.52\small{$\pm$0.17} & 0.52\small{$\pm$0.17} & 0.51\small{$\pm$0.15} & 0.66\small{$\pm$0.18} & 0.64\small{$\pm$0.16} & 0.52\small{$\pm$0.16} & 0.53\small{$\pm$0.19} & 0.62\small{$\pm$0.18} \\
					Random & 0.26\small{$\pm$0.13} & 0.26\small{$\pm$0.13} & 0.27\small{$\pm$0.13} & 0.47\small{$\pm$0.11} & 0.23\small{$\pm$0.13} & 0.28\small{$\pm$0.10} & 0.28\small{$\pm$0.11} & 0.24\small{$\pm$0.14} \\
					Fact  & 0.35\small{$\pm$0.08} & 0.36\small{$\pm$0.08} & 0.35\small{$\pm$0.09} & 0.48\small{$\pm$0.08} & 0.31\small{$\pm$0.10} & 0.35\small{$\pm$0.07} & 0.33\small{$\pm$0.09} & 0.29\small{$\pm$0.11} \\
					Matching & 0.53\small{$\pm$0.17} & 0.51\small{$\pm$0.18} & 0.52\small{$\pm$0.16} & 0.89\small{$\pm$0.08} & 0.58\small{$\pm$0.15} & 0.51\small{$\pm$0.16} & 0.55\small{$\pm$0.21} & 0.60\small{$\pm$0.21} \\
					DRM   & 0.81\small{$\pm$0.08} & 0.80\small{$\pm$0.08} & 0.80\small{$\pm$0.08} & 0.85\small{$\pm$0.06} & 0.77\small{$\pm$0.15} & 0.79\small{$\pm$0.09} & 0.81\small{$\pm$0.10} & 0.80\small{$\pm$0.08} \\
					\bottomrule
				\end{tabular}%
			}
			\label{tab:rank_cor_all}%
		\end{table}%
		\begin{figure}[h] 
			\centering
			\includegraphics[width=1\columnwidth]{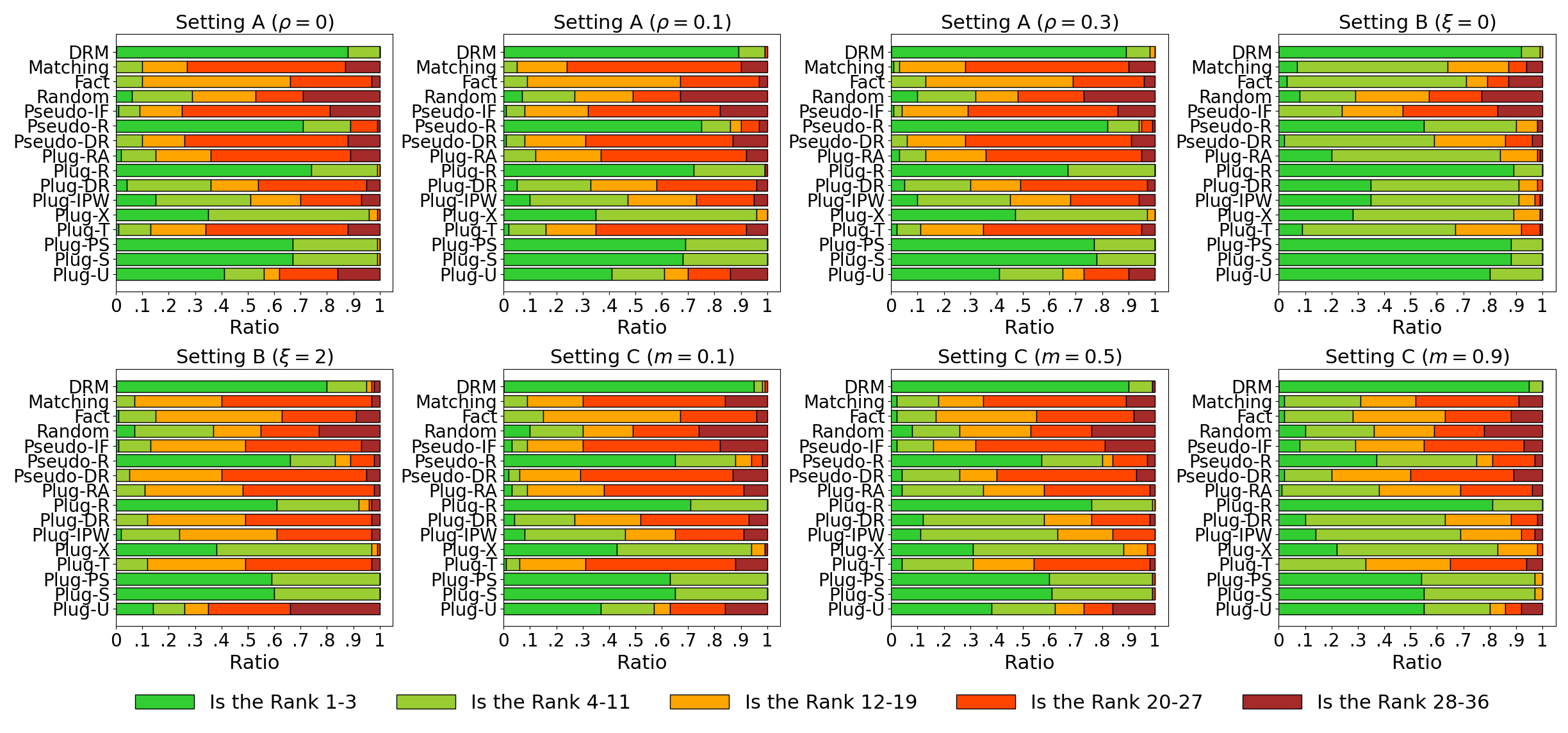} 
			\caption{The stacked bar chart showing the distribution of the selected estimator's rank for each evaluation metric across rank intervals: [1-3], [4-11], [12-19], [20-27], and [28-36]. The greener (or redder) color indicates that the selected estimator ranks higher (or lower). For example, the {\color{Brown}\textbf{dark red}} (or {\color{Green!100}\textbf{green}}) indicates the percentage of cases (out of 100 experiments) where the selected estimator ranks among the worst 9 estimators, specifically as ranks 28, 29, ..., or 36 (or among the best 3 estimators, specifically as ranks 1, 2, or 3).}
			\label{fig:stacked_bar_all}
		\end{figure}

			\paragraph{Regret comparison.} The results presented in Table \ref{tab:reg_all} demonstrate consistently good performance from the DRM selector across various settings. In setting A, the DRM selector outperforms other selectors as the CATE complexity ($\rho$) varies. Additionally, Plug-R, Plug-S, and Plug-PS also perform well in terms of the Regret criterion, which aligns with prior findings in \cite{schuler2018comparison} that the R-objective is excellent in many cases. Note that the strong performance of Plug-S and Plug-PS may be due to less pronounced heterogeneity in the CATE function compared to the outcome function in the data generating process. We also compare the PEHE performance (i.e., $\mathcal{R}^{oracle}(\hat{\tau}_{select})$) of different selectors in Table \ref{tab:pehe_all} of Section \ref{app:sec_all}. The results indicate that Plug-R, Plug-S, and Plug-PS tend to exhibit better PEHE as the CATE complexity decreases, aligning with the findings in \cite{pmlr-v202-curth23b}. In setting B, the DRM selector demonstrates robustness against selection bias (controlled by $\xi$) compared to many baselines. However, for the case $\xi=2$, DRM selects a poor estimator 1 or 2 times out of 100 experiments, as shown in Figure \ref{fig:stacked_bar_all}. Although this weakens its overall performance, DRM still outperforms many baselines in this scenario. In the scenario $\xi=0$ where no selection bias is present, the factual selection criterion performs better in this specific setting. In this case, DRM does not demonstrate a significant advantage, as there is no distribution shift caused by selection bias. In setting C where the unconfoundedness assumption is violated, most selectors exhibit inferior performance. In contrast, DRM demonstrates consistent outperformance across all three cases, and its superiority becomes particularly significant as $m$ increases to 0.9, showcasing its robustness against the distribution shift arising from unobserved confounders.
			
			\paragraph{Ranking ability.}  In Table \ref{tab:rank_cor_all}, the DRM method demonstrates favorable performance in ranking estimators, surpassing certain Plug- (e.g., U, T, IPW, DR, RA) and Pseudo- (e.g., DR, IF) selectors. In comparison to other nuisance-free baselines (Random, Fact, and Matching), DRM achieves significantly superior ranking ability. However, compared to Plug-S, -PS, -X, and -R, it does not exhibit remarkable performance in ranking CATE estimators, possibly due to the fact that DRM selects estimators based on their distributionally robust (worst-case) performance. Indeed, the definition of ranking inherently involves the concept of expected (average) performance, which is not determined solely by either the best or worst performance. While distributionally robust performance serves as a suitable criterion for selecting players to participate in the Olympics, it may not be a reasonable standard for ranking players' average performance. Therefore, it would be intriguing to explore some ways in future research that can enhance the ranking ability of our DRM selector.

			\paragraph{Variance analysis.} Table \ref{tab:reg_all} indicates that baseline selectors tend to exhibit higher variances in Regret performance. This is primarily due to the wide range of PEHE performances across the 36 CATE estimators. If a selector consistently selects either good or bad estimators, the variance would not be very large. To investigate this further, we sorted all 36 estimators in ascending order based on their $\mathcal{R}^{oracle}(\hat{\tau})$ values, resulting in the sorted list: $[\mathcal{R}^{oracle}(\hat{\tau}_1), \dots, \mathcal{R}^{oracle}(\hat{\tau}_J)]$. We then determine the actual rank of the selected estimator within this list and visualize the distribution of these 100 ranks using a stacked bar chart. Figure \ref{fig:stacked_bar_all} shows that many baseline methods tend to select CATE estimators from various percentile ranges, leading to high variance across the 100 selections. Notably, the DRM selector consistently chooses higher-ranked (i.e., better performing in PEHE) estimators, demonstrating its robustness in CATE estimator selection.
			
			\paragraph{Potential improvements.} There are several potential improvements based on the current experimental settings. First, the existing results suggest that Plug-S performs better than Plug-T, indicating that the complexity of CATE function is relatively simple. It would help to provide more comprehensive analysis if investigating how DRM compares to baselines when the CATE function is more complex. Second, since the impact of selection bias can vary with sample size \cite{alaa2018limits}, it is important to compare different selectors when the sample size is sufficiently large. Third, considering baselines that are specifically designed for addressing hidden confounders could provide valuable insights for testing different selectors under such conditions. We encourage deeper investigation of causal model selection without assuming unconfoundedness. Finally, it would be good if future studies will apply DRM and other selectors in Healthcare, Economics, and Business applications with real-world data, as CATE estimator selection plays an important role in personalized decision makings.

\section{Conclusion}
This paper sheds lights on the potential of robustness in CATE estimator selection. We propose a distributionally robust metric (DRM). The proposed metric is nuisance-free, eliminating the need to fit models for nuisance parameters (outcome function, propensity function, and plug-in learner). Additionally, it is well-targeted for selecting a robust CATE estimator. We provide a finite sample analysis that demonstrates the gap between $\hat{\mathcal{V}}^t(\hat{\tau})$ and $\mathcal{V}^t(\hat{\tau})$ reduces at a rate of $n^{-1/2}$ for $t \in \{0, 1\}$. The experimental results showcase that the CATE estimator selected by DRM demonstrate robustness to the distribution shift incurred by covariate shift and hidden confounders.

\paragraph{Limitations and future work.} This paper explores the potential of robustness in CATE estimator selection. However, we must acknowledge that our DRM method is not a silver bullet, as consistent estimation on the CATE are never attainable \citep{chernozhukov2018generic}. Here, we outline some challenges and suggest future research directions. First, while Proposition \ref{prop:kl_equal} provides useful guidance for setting ambiguity radius in the DRM algorithm, we cannot guarantee that the empirically-computed radius is optimal due to potential bias in the algorithm's approximation of KL-divergence. Second, as discussed in Section \ref{sec:exp_results}, enhancing the ranking capability of DRM is a promising area for further research. Moreover, our findings are based on KL-divergence. However, using other divergences, such as the Wasserstein distance, to construct the ambiguity set could incorporate more diverse distributions, despite the challenges in solving the dual formulation of the Wasserstein distributionally robust value. Simultaneously, exploring whether alternative divergences can yield a tighter bound for the PEHE error is also interesting \citep{pmlr-v80-alaa18a}. Finally, inspired by \cite{cui2024selective}, understanding how nuisance parameters influence metrics like plug-DR and pseudo-DR might be helpful in CATE estimator selection. We hope our methods and findings will spur interest in model selection for causal inference, as well as in related fields like domain adaptation and out-of-distribution generalization.


\section*{Acknowledgement}
Qi WU acknowledges the support from The CityU-JD Digits Joint Laboratory in Financial Technology and  Engineering, The Hong Kong Research Grants Council [General Research Fund 11219420/9043008], and The CityU APRC Grant 9610643. The work described in this paper was partially supported by the InnoHK initiative, the  Government of the HKSAR, and the Laboratory for AI-Powered Financial Technologies. We finally thank all the anonymous reviewers for their constructive suggestions.
\bibliography{neurips2024.bib}

\begin{thebibliography}{10}

\bibitem{abadie2023should}
Alberto Abadie, Susan Athey, Guido~W Imbens, and Jeffrey~M Wooldridge.
\newblock When should you adjust standard errors for clustering?
\newblock {\em The Quarterly Journal of Economics}, 138(1):1--35, 2023.

\bibitem{advani2019mostly}
Arun Advani, Toru Kitagawa, and Tymon S{\l}oczy{\'n}ski.
\newblock Mostly harmless simulations? using monte carlo studies for estimator
  selection.
\newblock {\em Journal of Applied Econometrics}, 34(6):893--910, 2019.

\bibitem{alaa2018limits}
Ahmed Alaa and Mihaela Schaar.
\newblock Limits of estimating heterogeneous treatment effects: Guidelines for
  practical algorithm design.
\newblock In {\em International Conference on Machine Learning}, pages
  129--138. PMLR, 2018.

\bibitem{pmlr-v80-alaa18a}
Ahmed Alaa and Mihaela van~der Schaar.
\newblock Limits of estimating heterogeneous treatment effects: Guidelines for
  practical algorithm design.
\newblock In Jennifer Dy and Andreas Krause, editors, {\em Proceedings of the
  35th International Conference on Machine Learning}, volume~80 of {\em
  Proceedings of Machine Learning Research}, pages 129--138. PMLR, 10--15 Jul
  2018.

\bibitem{alaa2019validating}
Ahmed Alaa and Mihaela Van Der~Schaar.
\newblock Validating causal inference models via influence functions.
\newblock In {\em International Conference on Machine Learning}, pages
  191--201. PMLR, 2019.

\bibitem{assaad2021counterfactual}
Serge Assaad, Shuxi Zeng, Chenyang Tao, Shounak Datta, Nikhil Mehta, Ricardo
  Henao, Fan Li, and Lawrence Carin.
\newblock Counterfactual representation learning with balancing weights.
\newblock In {\em International Conference on Artificial Intelligence and
  Statistics}, pages 1972--1980. PMLR, 2021.

\bibitem{athey2021using}
Susan Athey, Guido~W Imbens, Jonas Metzger, and Evan Munro.
\newblock Using wasserstein generative adversarial networks for the design of
  monte carlo simulations.
\newblock {\em Journal of Econometrics}, 2021.

\bibitem{athey2019generalized}
SUSAN ATHEY, JULIE TIBSHIRANI, and STEFAN WAGER.
\newblock Generalized random forests.
\newblock {\em The Annals of Statistics}, 47(2):1148--1178, 2019.

\bibitem{bica2021real}
Ioana Bica, Ahmed~M Alaa, Craig Lambert, and Mihaela Van Der~Schaar.
\newblock From real-world patient data to individualized treatment effects
  using machine learning: current and future methods to address underlying
  challenges.
\newblock {\em Clinical Pharmacology \& Therapeutics}, 109(1):87--100, 2021.

\bibitem{bica2022transfer}
Ioana Bica and Mihaela van~der Schaar.
\newblock Transfer learning on heterogeneous feature spaces for treatment
  effects estimation.
\newblock {\em Advances in Neural Information Processing Systems},
  35:37184--37198, 2022.

\bibitem{bottou2013counterfactual}
L{\'e}on Bottou, Jonas Peters, Joaquin Qui{\~n}onero-Candela, Denis~X Charles,
  D~Max Chickering, Elon Portugaly, Dipankar Ray, Patrice Simard, and
  Ed~Snelson.
\newblock Counterfactual reasoning and learning systems: The example of
  computational advertising.
\newblock {\em Journal of Machine Learning Research}, 14(11), 2013.

\bibitem{chen2016xgboost}
Tianqi Chen and Carlos Guestrin.
\newblock Xgboost: A scalable tree boosting system.
\newblock In {\em Proceedings of the 22nd acm sigkdd international conference
  on knowledge discovery and data mining}, pages 785--794, 2016.

\bibitem{chernozhukov2018double}
Victor Chernozhukov, Denis Chetverikov, Mert Demirer, Esther Duflo, Christian
  Hansen, Whitney Newey, and James Robins.
\newblock Double/debiased machine learning for treatment and structural
  parameters, 2018.

\bibitem{chernozhukov2018generic}
Victor Chernozhukov, Mert Demirer, Esther Duflo, and Ivan Fernandez-Val.
\newblock Generic machine learning inference on heterogeneous treatment effects
  in randomized experiments, with an application to immunization in india.
\newblock Technical report, National Bureau of Economic Research, 2018.

\bibitem{chu2021graph}
Zhixuan Chu, Stephen~L Rathbun, and Sheng Li.
\newblock Graph infomax adversarial learning for treatment effect estimation
  with networked observational data.
\newblock In {\em Proceedings of the 27th ACM SIGKDD Conference on Knowledge
  Discovery \& Data Mining}, pages 176--184, 2021.

\bibitem{cui2024selective}
Yifan Cui and EJ~Tchetgen~Tchetgen.
\newblock Selective machine learning of doubly robust functionals.
\newblock {\em Biometrika}, 111(2):517--535, 2024.

\bibitem{curth2021really}
Alicia Curth, David Svensson, Jim Weatherall, and Mihaela van~der Schaar.
\newblock Really doing great at estimating cate? a critical look at ml
  benchmarking practices in treatment effect estimation.
\newblock In {\em Thirty-fifth conference on neural information processing
  systems datasets and benchmarks track (round 2)}, 2021.

\bibitem{curth2021nonparametric}
Alicia Curth and Mihaela van~der Schaar.
\newblock Nonparametric estimation of heterogeneous treatment effects: From
  theory to learning algorithms.
\newblock In {\em International Conference on Artificial Intelligence and
  Statistics}, pages 1810--1818. PMLR, 2021.

\bibitem{curth2021inductive}
Alicia Curth and Mihaela van~der Schaar.
\newblock On inductive biases for heterogeneous treatment effect estimation.
\newblock {\em Advances in Neural Information Processing Systems},
  34:15883--15894, 2021.

\bibitem{pmlr-v202-curth23b}
Alicia Curth and Mihaela Van Der~Schaar.
\newblock In search of insights, not magic bullets: Towards demystification of
  the model selection dilemma in heterogeneous treatment effect estimation.
\newblock In {\em Proceedings of the 40th International Conference on Machine
  Learning}, volume 202 of {\em Proceedings of Machine Learning Research},
  pages 6623--6642. PMLR, 23--29 Jul 2023.

\bibitem{donnelly2021correction}
Robert Donnelly, David Blei, Susan Athey, et~al.
\newblock Correction to: Counterfactual inference for consumer choice across
  many product categories.
\newblock {\em Quantitative Marketing and Economics}, 19(3-4):409--409, 2021.

\bibitem{dorie2019automated}
Vincent Dorie, Jennifer Hill, Uri Shalit, Marc Scott, and Dan Cervone.
\newblock Automated versus do-it-yourself methods for causal inference: Lessons
  learned from a data analysis competition.
\newblock {\em Statistical Science}, 34(1):43--68, 2019.

\bibitem{farrell2015robust}
Max~H Farrell.
\newblock Robust inference on average treatment effects with possibly more
  covariates than observations.
\newblock {\em Journal of Econometrics}, 189(1):1--23, 2015.

\bibitem{fernandez2023comparison}
Carlos Fern{\'a}ndez-Lor{\'\i}a, Foster Provost, Jesse Anderton, Benjamin
  Carterette, and Praveen Chandar.
\newblock A comparison of methods for treatment assignment with an application
  to playlist generation.
\newblock {\em Information Systems Research}, 34(2):786--803, 2023.

\bibitem{fisherinverse}
Aaron Fisher.
\newblock Inverse-variance weighting for estimation of heterogeneous treatment
  effects.
\newblock In {\em Forty-first International Conference on Machine Learning}.

\bibitem{foster2023orthogonal}
Dylan~J Foster and Vasilis Syrgkanis.
\newblock Orthogonal statistical learning.
\newblock {\em The Annals of Statistics}, 51(3):879--908, 2023.

\bibitem{foster2011subgroup}
Jared~C Foster, Jeremy~MG Taylor, and Stephen~J Ruberg.
\newblock Subgroup identification from randomized clinical trial data.
\newblock {\em Statistics in medicine}, 30(24):2867--2880, 2011.

\bibitem{guo2020survey}
Ruocheng Guo, Lu~Cheng, Jundong Li, P~Richard Hahn, and Huan Liu.
\newblock A survey of learning causality with data: Problems and methods.
\newblock {\em ACM Computing Surveys (CSUR)}, 53(4):1--37, 2020.

\bibitem{guo2021ignite}
Ruocheng Guo, Jundong Li, Yichuan Li, K~Sel{\c{c}}uk Candan, Adrienne Raglin,
  and Huan Liu.
\newblock Ignite: A minimax game toward learning individual treatment effects
  from networked observational data.
\newblock In {\em Proceedings of the Twenty-Ninth International Conference on
  International Joint Conferences on Artificial Intelligence}, pages
  4534--4540, 2021.

\bibitem{hahn2020bayesian}
P~Richard Hahn, Jared~S Murray, and Carlos~M Carvalho.
\newblock Bayesian regression tree models for causal inference: Regularization,
  confounding, and heterogeneous effects (with discussion).
\newblock {\em Bayesian Analysis}, 15(3):965--1056, 2020.

\bibitem{hassanpour2019learning}
Negar Hassanpour and Russell Greiner.
\newblock Learning disentangled representations for counterfactual regression.
\newblock In {\em International Conference on Learning Representations}, 2019.

\bibitem{hill2011bayesian}
Jennifer~L Hill.
\newblock Bayesian nonparametric modeling for causal inference.
\newblock {\em Journal of Computational and Graphical Statistics},
  20(1):217--240, 2011.

\bibitem{holland1986statistics}
Paul~W Holland.
\newblock Statistics and causal inference.
\newblock {\em Journal of the American statistical Association},
  81(396):945--960, 1986.

\bibitem{hu2013kullback}
Zhaolin Hu and L~Jeff Hong.
\newblock Kullback-leibler divergence constrained distributionally robust
  optimization.
\newblock {\em Available at Optimization Online}, 1(2):9, 2013.

\bibitem{huang2023towards}
Yiyan Huang, Cheuk~Hang Leung, Shumin Ma, Zhiri Yuan, Qi~Wu, Siyi Wang,
  Dongdong Wang, and Zhixiang Huang.
\newblock Towards balanced representation learning for credit policy
  evaluation.
\newblock In {\em International Conference on Artificial Intelligence and
  Statistics}, pages 3677--3692. PMLR, 2023.

\bibitem{huang2022robust}
Yiyan Huang, Cheuk~Hang Leung, Qi~Wu, Xing Yan, Shumin Ma, Zhiri Yuan, Dongdong
  Wang, and Zhixiang Huang.
\newblock Robust causal learning for the estimation of average treatment
  effects.
\newblock In {\em 2022 International Joint Conference on Neural Networks
  (IJCNN)}, pages 1--9. IEEE, 2022.

\bibitem{huang2021causal}
Yiyan Huang, Cheuk~Hang Leung, Xing Yan, Qi~Wu, Nanbo Peng, Dongdong Wang, and
  Zhixiang Huang.
\newblock The causal learning of retail delinquency.
\newblock In {\em Proceedings of the AAAI Conference on Artificial
  Intelligence}, volume~35, pages 204--212, 2021.

\bibitem{huang2024dignet}
Yiyan Huang, WANG Siyi, Cheuk~Hang Leung, WU~Qi, WANG Dongdong, and Zhixiang
  Huang.
\newblock Dignet: Learning decomposed patterns in representation balancing for
  treatment effect estimation.
\newblock {\em Transactions on Machine Learning Research}, 2024.

\bibitem{johansson2016learning}
Fredrik Johansson, Uri Shalit, and David Sontag.
\newblock Learning representations for counterfactual inference.
\newblock In {\em International conference on machine learning}, pages
  3020--3029. PMLR, 2016.

\bibitem{johansson2022generalization}
Fredrik~D Johansson, Uri Shalit, Nathan Kallus, and David Sontag.
\newblock Generalization bounds and representation learning for estimation of
  potential outcomes and causal effects.
\newblock {\em The Journal of Machine Learning Research}, 23(1):7489--7538,
  2022.

\bibitem{kennedy2023towards}
Edward~H Kennedy.
\newblock Towards optimal doubly robust estimation of heterogeneous causal
  effects.
\newblock {\em Electronic Journal of Statistics}, 17(2):3008--3049, 2023.

\bibitem{kinyanjui2022adcb}
Newton~Mwai Kinyanjui and Fredrik~D Johansson.
\newblock Adcb: An alzheimer’s disease simulator for benchmarking
  observational estimators of causal effects.
\newblock In {\em Conference on Health, Inference, and Learning}, pages
  103--118. PMLR, 2022.

\bibitem{kitagawa2018should}
Toru Kitagawa and Aleksey Tetenov.
\newblock Who should be treated? empirical welfare maximization methods for
  treatment choice.
\newblock {\em Econometrica}, 86(2):591--616, 2018.

\bibitem{kuang2017treatment}
Kun Kuang, Peng Cui, Bo~Li, Meng Jiang, Shiqiang Yang, and Fei Wang.
\newblock Treatment effect estimation with data-driven variable decomposition.
\newblock In {\em Proceedings of the AAAI Conference on Artificial
  Intelligence}, volume~31, 2017.

\bibitem{kuang2020data}
Kun Kuang, Peng Cui, Hao Zou, Bo~Li, Jianrong Tao, Fei Wu, and Shiqiang Yang.
\newblock Data-driven variable decomposition for treatment effect estimation.
\newblock {\em IEEE Transactions on Knowledge and Data Engineering},
  34(5):2120--2134, 2020.

\bibitem{kuhn2019wasserstein}
Daniel Kuhn, Peyman~Mohajerin Esfahani, Viet~Anh Nguyen, and Soroosh
  Shafieezadeh-Abadeh.
\newblock Wasserstein distributionally robust optimization: Theory and
  applications in machine learning.
\newblock In {\em Operations research \& management science in the age of
  analytics}, pages 130--166. Informs, 2019.

\bibitem{kunzel2019metalearners}
S{\"o}ren~R K{\"u}nzel, Jasjeet~S Sekhon, Peter~J Bickel, and Bin Yu.
\newblock Metalearners for estimating heterogeneous treatment effects using
  machine learning.
\newblock {\em Proceedings of the national academy of sciences},
  116(10):4156--4165, 2019.

\bibitem{levy2020large}
Daniel Levy, Yair Carmon, John~C Duchi, and Aaron Sidford.
\newblock Large-scale methods for distributionally robust optimization.
\newblock {\em Advances in Neural Information Processing Systems},
  33:8847--8860, 2020.

\bibitem{li2022random}
Shuangning Li and Stefan Wager.
\newblock Random graph asymptotics for treatment effect estimation under
  network interference.
\newblock {\em The Annals of Statistics}, 50(4):2334--2358, 2022.

\bibitem{li2024causal}
Yijun Li, Cheuk~Hang Leung, Xiangqian Sun, Chaoqun Wang, Yiyan Huang, Xing Yan,
  Qi~Wu, Dongdong Wang, and Zhixiang Huang.
\newblock The causal impact of credit lines on spending distributions.
\newblock In {\em Proceedings of the AAAI Conference on Artificial
  Intelligence}, volume~38, pages 180--187, 2024.

\bibitem{louizos2017causal}
Christos Louizos, Uri Shalit, Joris~M Mooij, David Sontag, Richard Zemel, and
  Max Welling.
\newblock Causal effect inference with deep latent-variable models.
\newblock {\em Advances in neural information processing systems}, 30, 2017.

\bibitem{ma2020understanding}
Shumin Ma, Cheuk~Hang Leung, Qi~Wu, Wei Liu, and Nanbo Peng.
\newblock Understanding distributional ambiguity via non-robust chance
  constraint.
\newblock In {\em Proceedings of the First ACM International Conference on AI
  in Finance}, pages 1--8, 2020.

\bibitem{mahajan2022empirical}
Divyat Mahajan, Ioannis Mitliagkas, Brady Neal, and Vasilis Syrgkanis.
\newblock Empirical analysis of model selection for heterogenous causal effect
  estimation.
\newblock {\em International Conference on Learning Representations}, 2024.

\bibitem{mohajerin2018data}
Peyman Mohajerin~Esfahani and Daniel Kuhn.
\newblock Data-driven distributionally robust optimization using the
  wasserstein metric: performance guarantees and tractable reformulations.
\newblock {\em Mathematical Programming}, 171(1-2):115--166, 2018.

\bibitem{nie2021quasi}
Xinkun Nie and Stefan Wager.
\newblock Quasi-oracle estimation of heterogeneous treatment effects.
\newblock {\em Biometrika}, 108(2):299--319, 2021.

\bibitem{nogueira2022methods}
Ana~Rita Nogueira, Andrea Pugnana, Salvatore Ruggieri, Dino Pedreschi, and
  Jo{\~a}o Gama.
\newblock Methods and tools for causal discovery and causal inference.
\newblock {\em Wiley interdisciplinary reviews: data mining and knowledge
  discovery}, 12(2):e1449, 2022.

\bibitem{noh2014bias}
Yung-Kyun Noh, Masashi Sugiyama, Song Liu, Marthinus~C Plessis, Frank~Chongwoo
  Park, and Daniel~D Lee.
\newblock Bias reduction and metric learning for nearest-neighbor estimation of
  kullback-leibler divergence.
\newblock In {\em Artificial Intelligence and Statistics}, pages 669--677.
  PMLR, 2014.

\bibitem{oprescu2019orthogonal}
Miruna Oprescu, Vasilis Syrgkanis, and Zhiwei~Steven Wu.
\newblock Orthogonal random forest for causal inference.
\newblock In {\em International Conference on Machine Learning}, pages
  4932--4941. PMLR, 2019.

\bibitem{parikh2022validating}
Harsh Parikh, Carlos Varjao, Louise Xu, and Eric~Tchetgen Tchetgen.
\newblock Validating causal inference methods.
\newblock In {\em International Conference on Machine Learning}, pages
  17346--17358. PMLR, 2022.

\bibitem{pflug2023multistage}
Georg~Ch Pflug.
\newblock Multistage stochastic decision problems: Approximation by recursive
  structures and ambiguity modeling.
\newblock {\em European Journal of Operational Research}, 306(3):1027--1039,
  2023.

\bibitem{qian2021synctwin}
Zhaozhi Qian, Yao Zhang, Ioana Bica, Angela Wood, and Mihaela van~der Schaar.
\newblock Synctwin: Treatment effect estimation with longitudinal outcomes.
\newblock {\em Advances in Neural Information Processing Systems},
  34:3178--3190, 2021.

\bibitem{rolling2014model}
Craig~A Rolling and Yuhong Yang.
\newblock Model selection for estimating treatment effects.
\newblock {\em Journal of the Royal Statistical Society Series B: Statistical
  Methodology}, 76(4):749--769, 2014.

\bibitem{rosenbaum1983central}
Paul~R Rosenbaum and Donald~B Rubin.
\newblock The central role of the propensity score in observational studies for
  causal effects.
\newblock {\em Biometrika}, 70(1):41--55, 1983.

\bibitem{rubin2005causal}
Donald~B Rubin.
\newblock Causal inference using potential outcomes: Design, modeling,
  decisions.
\newblock {\em Journal of the American Statistical Association},
  100(469):322--331, 2005.

\bibitem{saito2020counterfactual}
Yuta Saito and Shota Yasui.
\newblock Counterfactual cross-validation: Stable model selection procedure for
  causal inference models.
\newblock In {\em International Conference on Machine Learning}, pages
  8398--8407. PMLR, 2020.

\bibitem{schuler2018comparison}
Alejandro Schuler, Michael Baiocchi, Robert Tibshirani, and Nigam Shah.
\newblock A comparison of methods for model selection when estimating
  individual treatment effects.
\newblock {\em arXiv preprint arXiv:1804.05146}, 2018.

\bibitem{schuler2017synth}
Alejandro Schuler, Ken Jung, Robert Tibshirani, Trevor Hastie, and Nigam Shah.
\newblock Synth-validation: Selecting the best causal inference method for a
  given dataset.
\newblock {\em arXiv preprint arXiv:1711.00083}, 2017.

\bibitem{shalit2017estimating}
Uri Shalit, Fredrik~D Johansson, and David Sontag.
\newblock Estimating individual treatment effect: generalization bounds and
  algorithms.
\newblock In {\em International conference on machine learning}, pages
  3076--3085. PMLR, 2017.

\bibitem{shi2019adapting}
Claudia Shi, David Blei, and Victor Veitch.
\newblock Adapting neural networks for the estimation of treatment effects.
\newblock {\em Advances in neural information processing systems}, 32, 2019.

\bibitem{wager2018estimation}
Stefan Wager and Susan Athey.
\newblock Estimation and inference of heterogeneous treatment effects using
  random forests.
\newblock {\em Journal of the American Statistical Association},
  113(523):1228--1242, 2018.

\bibitem{wang2021flaml}
Chi Wang, Qingyun Wu, Markus Weimer, and Erkang Zhu.
\newblock Flaml: A fast and lightweight automl library.
\newblock {\em Proceedings of Machine Learning and Systems}, 3:434--447, 2021.

\bibitem{wang2021sinkhorn}
Jie Wang, Rui Gao, and Yao Xie.
\newblock Sinkhorn distributionally robust optimization.
\newblock {\em arXiv preprint arXiv:2109.11926}, 2021.

\bibitem{wang2006nearest}
Qing Wang, Sanjeev~R Kulkarni, and Sergio Verd{\'u}.
\newblock A nearest-neighbor approach to estimating divergence between
  continuous random vectors.
\newblock In {\em 2006 IEEE International Symposium on Information Theory},
  pages 242--246. IEEE, 2006.

\bibitem{wu2022learning}
Anpeng Wu, Junkun Yuan, Kun Kuang, Bo~Li, Runze Wu, Qiang Zhu, Yueting Zhuang,
  and Fei Wu.
\newblock Learning decomposed representations for treatment effect estimation.
\newblock {\em IEEE Transactions on Knowledge and Data Engineering},
  35(5):4989--5001, 2022.

\bibitem{yao2021survey}
Liuyi Yao, Zhixuan Chu, Sheng Li, Yaliang Li, Jing Gao, and Aidong Zhang.
\newblock A survey on causal inference.
\newblock {\em ACM Transactions on Knowledge Discovery from Data (TKDD)},
  15(5):1--46, 2021.

\bibitem{yao2018representation}
Liuyi Yao, Sheng Li, Yaliang Li, Mengdi Huai, Jing Gao, and Aidong Zhang.
\newblock Representation learning for treatment effect estimation from
  observational data.
\newblock {\em Advances in neural information processing systems}, 31, 2018.

\bibitem{yoon2018ganite}
Jinsung Yoon, James Jordon, and Mihaela Van Der~Schaar.
\newblock Ganite: Estimation of individualized treatment effects using
  generative adversarial nets.
\newblock In {\em International conference on learning representations}, 2018.

\bibitem{zhang2019medical}
Linying Zhang, Yixin Wang, Anna Ostropolets, Jami~J Mulgrave, David~M Blei, and
  George Hripcsak.
\newblock The medical deconfounder: assessing treatment effects with electronic
  health records.
\newblock In {\em Machine Learning for Healthcare Conference}, pages 490--512.
  PMLR, 2019.

\bibitem{zhang2020learning}
Yao Zhang, Alexis Bellot, and Mihaela Schaar.
\newblock Learning overlapping representations for the estimation of
  individualized treatment effects.
\newblock In {\em International Conference on Artificial Intelligence and
  Statistics}, pages 1005--1014. PMLR, 2020.

\end{thebibliography}
\bibliographystyle{plain}


\appendix

\clearpage
\section*{Appendix}
\section{CATE Estimation Strategies}
\subsection{CATE Learners} \label{sec:CATE Learners}
We now detail how to construct CATE learners using the observed samples $\{(X_{i},T_{i},Y_{i})\}_{i=1}^{n}$. Note that CATE learners are learned on the training set, so the sample size $n$ here equals the training sample size. Denote $n_t$ by the sample size in the treat group, and $n_c$ by the sample size in the control group such that $n=n_t+n_c$.
\begin{itemize}
	\item S-learner: Let predictors=$(X,T)$, response=$Y$. Train a model $\hat{\mu}(X,T)$. Then we obtain $\hat{\tau}_{S}(X)$:
	\begin{align*}
		\hat{\tau}_{S}(X)=\hat{\mu}(X,1) - \hat{\mu}(X,0).
	\end{align*}
	\item T-learner: Let predictors=$X^T$ (covariates in the treat), response=$Y^T$ (outcome in the treat). Train a model $\hat{\mu}_1(X)$. Let predictors=$X^C$ (covariates in the control), response=$Y^C$ (outcome in the control). Train a model $\hat{\mu}_0(X)$. Then we obtain $\hat{\tau}_{T}(X)$:
	\begin{align*}
		\hat{\tau}_{T}(X)=\hat{\mu}_1(X) - \hat{\mu}_0(X).
	\end{align*}
	\item PS-learner: Fisrt-step: Train $\hat{\tau}_{S}(X)$ using the above-mentioned step in S-learner. Second-step: Let predictors=$X$, response=$\hat{\tau}_{S}(X)$. Train a model $\hat{\tau}_{PS}(X)$ from the following objective:
	\begin{align*}
		\hat{\tau}_{PS}=\mathop{\arg\min}_{\tau} \; \frac{1}{n}\sum_{i=1}^{n}(\tau(X_i) - \hat{\tau}_{S}(X_i))^2.
	\end{align*}
	\item IPW-learner: First-step: let predictors=$X$, response=$T$. Train a propensity score model $\hat{\pi}(X)$. Construct surrogate of CATE using pseudo-outcomes with inverse propensity weighting (IPW) formula: $Y_{IPW}^{1,0}=Y_{IPW}^{1}-Y_{IPW}^{0}$, where $Y_{IPW}^{1}=\frac{TY}{\hat{\pi}(X)}$ and $Y_{IPW}^{0}=\frac{(1-T)Y}{1-\hat{\pi}(X)}$. Train a model $\hat{\tau}_{IPW}(X)$ from the following objective:
	\begin{align*}
		\hat{\tau}_{IPW}=\mathop{\arg\min}_{\tau} \; \frac{1}{n}\sum_{i=1}^{n}(\tau(X_i) - Y_{i,IPW}^{1,0})^2.
	\end{align*}
	\item X-learner \citep{kunzel2019metalearners}: First-step: Train $\hat{\mu}_1(X)$ and $\hat{\mu}_0(X)$ using the the above-mentioned procedure in T-learner. Train a propensity score model $\hat{\pi}(X)$ using the the above-mentioned procedure in IPW-learner. Second-step: Let predictors=$X^T$, response=$\hat{\mu}_1(X^T)-Y^T$, and predictors=$X^C$, response=$\hat{\mu}_0(X^C)-Y^C$. Obtain a model $\hat{\tau}_{X}(X)$ by learning two separate functions $\hat{\tau}^1_{X}(X)$ and $\hat{\tau}^0_{X}(X)$:
	\begin{align*}
		&\hat{\tau}_{X}(X) = (1-\hat{\pi}(X))\hat{\tau}^1_{X}(X) + \hat{\pi}(X)\hat{\tau}^0_{X}(X),\\
		&\hat{\tau}^1_{X}=\mathop{\arg\min}_{\tau} \; \frac{1}{n_t}\sum_{i=1}^{n_t}(\tau(X_i) - (Y_i-\hat{\mu}_0(X_i)))^2,\\
		&\hat{\tau}^0_{X}=\mathop{\arg\min}_{\tau} \; \frac{1}{n_c}\sum_{i=1}^{n_c}(\tau(X_i) - (\hat{\mu}_1(X_i)-Y_i))^2.
	\end{align*}
	\item U-learner \citep{fisherinverse, nie2021quasi}: First-step: Let predictors=$X$, response=$Y$. Train a model $\hat{\mu}(X)$ to approximate the conditional mean outcome $\mathbb{E}[Y|X]$. Train a propensity score model $\hat{\pi}(X)$ using the the above-mentioned procedure in IPW-learner. Second-step: Compute the outcome residual $\xi = Y-\hat{\mu}(X)$ and treatment residual $\nu=T-\hat{\pi}(X)$. Train a model $\hat{\tau}_{U}(X)$ from the following objective:
	\begin{align*}
		\hat{\tau}_{U}=\mathop{\arg\min}_{\tau} \; \frac{1}{n}\sum_{i=1}^{n}(\frac{\xi_i}{\nu_i}- \tau(X_i))^2.
	\end{align*}
	\item DR-learner \citep{kennedy2023towards, foster2023orthogonal}: First-step: Train $\hat{\mu}_1(X)$ and $\hat{\mu}_0(X)$ using the the above-mentioned procedure in T-learner. Train a propensity score model $\hat{\pi}(X)$ using the the above-mentioned procedure in IPW-learner. Second-step: Construct surrogate of CATE using pseudo-outcomes with doubly robust (DR) formula: $Y_{DR}^{1,0}=Y_{DR}^{1}-Y_{DR}^{0}$, where $Y_{DR}^{1}=\hat{\mu}_1(X)+\frac{T}{\hat{\pi}(X)}(Y-\hat{\mu}_1(X))$ and $Y_{DR}^{0}=\hat{\mu}_0(X)+\frac{1-T}{1-\hat{\pi}(X)}(Y-\hat{\mu}_0(X))$. Train a model $\hat{\tau}_{DR}(X)$ from the following objective:
	\begin{align*}
		\hat{\tau}_{DR}=\mathop{\arg\min}_{\tau} \; \frac{1}{n}\sum_{i=1}^{n}(\tau(X_i) - Y_{i,DR}^{1,0})^2.
	\end{align*}
	\item R-learner \citep{nie2021quasi}: First-step: Let predictors=$X$, response=$Y$. Train a model $\hat{\mu}(X)$ to approximate the conditional mean outcome $\mathbb{E}[Y|X]$. Train a propensity score model $\hat{\pi}(X)$ using the the above-mentioned procedure in IPW-learner. Second-step: Compute the outcome residual $\xi = Y-\hat{\mu}(X)$ and treatment residual $\nu=T-\hat{\pi}(X)$. Train a model $\hat{\tau}_{R}(X)$ from the following objective:
	\begin{align*}
		\hat{\tau}_{R}=\mathop{\arg\min}_{\tau} \; \frac{1}{n}\sum_{i=1}^{n}(\xi_i - \nu_i\tau(X_i))^2.
	\end{align*}
	\item RA-learner \citep{curth2021nonparametric}: First-step: Train $\hat{\mu}_1(X)$ and $\hat{\mu}_0(X)$ using the the above-mentioned procedure in T-learner. Second-step: Construct surrogate of CATE using pseudo-outcomes with regression adjustment (RA) formula: $Y_{RA}=T(Y-\hat{\mu}_0(X))+(1-T)(\hat{\mu}_1(X)-Y)$. Train a model $\hat{\tau}_{RA}(X)$ from the following objective:
	\begin{align*}
		\hat{\tau}_{RA}=\mathop{\arg\min}_{\tau} \; \frac{1}{n}\sum_{i=1}^{n}(\tau(X_i)-Y_{i,RA})^2.
	\end{align*}
\end{itemize}

\subsection{CATE Selectors} \label{sec:CATE selectors}
We now detail how to construct CATE selectors using the observed samples $\{(X_{i},T_{i},Y_{i})\}_{i=1}^{n}$. Note that CATE selectors are constructed on the validation set, so the sample size $n$ here equals the validation sample size.
\begin{itemize}
	\item Plug-in selector: Obtain any CATE learners $\tilde{\tau}$ using the observational validation data. Then plug-in $\tilde{\tau}$ into the following metric $\mathcal{R}_{\tilde{\tau}}^{plug}(\hat{\tau})$:
	\begin{align*}
		\mathcal{R}_{\tilde{\tau}}^{plug}(\hat{\tau})=\sqrt{\frac{1}{n} \sum_{i=1}^{n} (\hat{\tau}(X_i) - \tilde{\tau}(X_i))^2}.
	\end{align*}
	For each plug-in selector $\tilde{\tau}$, the selected $j^*$-th CATE estimator is $\hat{\tau}_{j^*}$, where $j^*= \mathop{\arg \min}_{j \in \{1, \dots, J\}}\mathcal{R}_{\tilde{\tau}}^{plug}(\hat{\tau}_j)$.
	\item Pseudo-outcome selector: 
	\begin{enumerate}
		\item Pseudo-DR: Utilize validation data to estimate nuisance parameters $(\tilde{\mu}_1, \tilde{\mu}_0, \tilde{\pi})$, following the procedure described in Section \ref{sec:CATE Learners}. $\tilde{Y}_{DR}=\tilde{Y}_{DR}^{1}-\tilde{Y}_{DR}^{0}$, where $\tilde{Y}_{DR}^{1}=\tilde{\mu}_1(X)+\frac{T}{\tilde{\pi}(X)}(Y-\tilde{\mu}_1(X))$ and $\tilde{Y}_{DR}^{0}=\tilde{\mu}_0(X)+\frac{1-T}{1-\tilde{\pi}(X)}(Y-\tilde{\mu}_0(X))$. Then the pseudo-DR metric is
		\begin{align*}
			\mathcal{R}_{DR}^{pseudo}(\hat{\tau})=\sqrt{\frac{1}{n} \sum_{i=1}^{n} (\hat{\tau}(X_i) - \tilde{Y}_{i, DR})^2}.
		\end{align*}
		For pseudo-DR selector, the selected $j^*$-th CATE estimator is $\hat{\tau}_{j^*}$, where $j^*= \mathop{\arg \min}_{j \in \{1, \dots, J\}}\mathcal{R}_{DR}^{pseudo}(\hat{\tau}_j)$.
		\item Pseudo-R: Utilize validation data to estimate nuisance parameters $(\tilde{\mu}, \tilde{\pi})$, following the procedure described in Section \ref{sec:CATE Learners}. Then the pseudo-R metric is
		\begin{align*}
			\mathcal{R}_{R}^{pseudo}(\hat{\tau})=\sqrt{\frac{1}{n} \sum_{i=1}^{n} ( (Y_i-\tilde{\mu}(X_i)) - \hat{\tau}(X_i)(T_i-\tilde{\pi}(X_i)) )^2}.
		\end{align*}
		For pseudo-R selector, the selected $j^*$-th CATE estimator is $\hat{\tau}_{j^*}$, where $j^*= \mathop{\arg \min}_{j \in \{1, \dots, J\}}\mathcal{R}_{R}^{pseudo}(\hat{\tau}_j)$.
		\item Pseudo-IF \citep{alaa2019validating}: Utilize validation data to estimate nuisance parameters $(\tilde{\mu}_1, \tilde{\mu}_0, \tilde{\pi})$, following the procedure described in Section \ref{sec:CATE Learners}. Let $\tilde{\tau}(X)=(\tilde{\mu}_1(X)-\tilde{\mu}_0(X))$. Then the pseudo-IF metric is
		\begin{align*}
			&\mathcal{R}_{IF}^{pseudo}(\hat{\tau})=\sqrt{\frac{1}{n} \sum_{i=1}^{n} \left( (1-B_i)\tilde{\tau}^2(X_i) +B_iY_i(\tilde{\tau}(X_i)-\hat{\tau}(X_i)) - A_i(\tilde{\tau}(X_i)-\hat{\tau}(X_i))^2 + \hat{\tau}^2(X_i)\right)},\\
			& where \ A_i=T_i-\tilde{\pi}(X_i), \ B_i=2T_i(T_i-\tilde{\pi}(X_i))C_i^{-1}, \ C_i=\tilde{\pi}(X_i)(1-\tilde{\pi}(X_i)).
		\end{align*}
		For pseudo-IF selector, the selected $j^*$-th CATE estimator is $\hat{\tau}_{j^*}$, where $j^*= \mathop{\arg \min}_{j \in \{1, \dots, J\}}\mathcal{R}_{IF}^{pseudo}(\hat{\tau}_j)$.
		\item Other pseudo-outcome selector: By manipulating the formula of $\tilde{Y}$, it is possible to create additional pseudo-outcome selectors, such as the pseudo-IPW selector. In our paper, we choose pseudo-DR as the baseline because it is representative in the causal inference literature and it often demonstrates superior performance, owing to its doubly robust property.
	\end{enumerate}
\end{itemize}

\section{Proofs}
\subsection{Proof of Proposition \ref{prop:pehe_decompose}} \label{app:proof_pehe_decompose}
\begin{proof}
	\begin{equation*}
		\begin{aligned}
			&\mathbb{E}[(\hat{\tau}(X)-\tau_{true}(X))^2]\\
			&= \mathbb{E}[(\hat{\tau}(X)-(\mu_1(X)-\mu_0(X)))^2] \\
			&= \mathbb{E}[(\hat{\tau}(X)-\mu_1(X)+\mu_0(X))^2] \\
			&=\mathbb{E}[(\hat{\tau}(X)-\mu_1(X))^2] + \mathbb{E}[\mu_0(X)^2] + 2 \mathbb{E}[(\hat{\tau}(X)-\mu_1(X))\mu_0(X)] \\
			&=\mathbb{E}[\hat{\tau}(X)^2] + \mathbb{E}[\mu_1(X)^2] - 2 \mathbb{E}[\hat{\tau}(X)\mu_1(X)] + \mathbb{E}[\mu_0(X)^2] + 2 \mathbb{E}[\hat{\tau}(X)\mu_0(X)] - 2\mathbb{E}[\mu_1(X)\mu_0(X)]\\
			&=\mathbb{E}[\hat{\tau}(X)^2] - 2 \mathbb{E}[\hat{\tau}(X)(\mu_1(X)-Y^1+Y^1)] + 2 \mathbb{E}[\hat{\tau}(X)(\mu_0(X)-Y^0+Y^0)]\\
			& \quad  + \mathbb{E}[\mu_1(X)^2] + \mathbb{E}[\mu_0(X)^2]- 2\mathbb{E}[\mu_1(X)\mu_0(X)]\\
			&=\mathbb{E}[\hat{\tau}(X)^2] - 2 \mathbb{E}[\hat{\tau}(X)Y^1] - 2 \mathbb{E}[\hat{\tau}(X)(\mu_1(X)-Y^1)] + 2 \mathbb{E}[\hat{\tau}(X)Y^0] + 2 \mathbb{E}[\hat{\tau}(X)(\mu_0(X)-Y^0)]\\
			& \quad  + \mathbb{E}[\mu_1(X)^2] + \mathbb{E}[\mu_0(X)^2]- 2\mathbb{E}[\mu_1(X)\mu_0(X)]\\
			&=\mathbb{E}[\hat{\tau}(X)^2] - 2 \mathbb{E}[\hat{\tau}(X)Y^1] - 2 \mathbb{E}[\mathbb{E}[\hat{\tau}(X)\mu_1(X)-\hat{\tau}(X)Y^1|X]] + 2 \mathbb{E}[\hat{\tau}(X)Y^0] \\
			& \quad + 2 \mathbb{E}[\mathbb{E}[\hat{\tau}(X)\mu_0(X)-\hat{\tau}(X)Y^0|X]] 
			+ \mathbb{E}[\mu_1(X)^2] + \mathbb{E}[\mu_0(X)^2]- 2\mathbb{E}[\mu_1(X)\mu_0(X)]\\
			&=\mathbb{E}[\hat{\tau}(X)^2] - 2 \mathbb{E}[\hat{\tau}(X)Y^1] - 2 \mathbb{E}[\hat{\tau}(X)\mu_1(X)-\hat{\tau}(X)\mathbb{E}[Y^1|X]] + 2 \mathbb{E}[\hat{\tau}(X)Y^0] \\
			& \quad + 2 \mathbb{E}[\hat{\tau}(X)\mu_0(X)-\hat{\tau}(X)\mathbb{E}[Y^0|X]] 
			+ \mathbb{E}[\mu_1(X)^2] + \mathbb{E}[\mu_0(X)^2]- 2\mathbb{E}[\mu_1(X)\mu_0(X)]\\
			&=\mathbb{E}[\hat{\tau}(X)^2] - 2 \mathbb{E}[\hat{\tau}(X)Y^1] - 2 \mathbb{E}[\hat{\tau}(X)\mu_1(X)-\hat{\tau}(X)\mu_1(X)] + 2 \mathbb{E}[\hat{\tau}(X)Y^0]\\ 
			& \quad + 2 \mathbb{E}[\hat{\tau}(X)\mu_0(X)-\hat{\tau}(X)\mu_0(X)] 
			+ \mathbb{E}[\mu_1(X)^2] + \mathbb{E}[\mu_0(X)^2]- 2\mathbb{E}[\mu_1(X)\mu_0(X)]\\
			&=\mathbb{E}[\hat{\tau}(X)^2] + 2 \mathbb{E}[\hat{\tau}(X)Y^0] - 2 \mathbb{E}[\hat{\tau}(X)Y^1] + \mathbb{E}[\mu_1(X)^2] + \mathbb{E}[\mu_0(X)^2]- 2\mathbb{E}[\mu_1(X)\mu_0(X)]\\
			&= \mathbb{E}[\hat{\tau}(X)^2] + 2 \mathbb{E}[\hat{\tau}(X)Y^0] - 2 \mathbb{E}[\hat{\tau}(X)Y^1] + \zeta.
		\end{aligned}
	\end{equation*}
\end{proof}

\subsection{Proof of Proposition \ref{prop:kl_equal}} \label{app:proof_kl_equal}
The following Proposition \ref{prop:prob_decompose} is useful in proving Proposition \ref{prop:kl_equal}.
\begin{proposition} \label{prop:prob_decompose} Assuming the random variable tuple $(X, T, Y^1, Y^0)$ satisfies Assumption \ref{assump}, we have
	\begin{equation}
		\begin{aligned}
			&p(X,Y^0, Y^1|T=0)=	p(Y^0, Y^1|X)p(X|T=0);\\
			&p(X,Y^0, Y^1|T=1)=	p(Y^0, Y^1|X)p(X|T=1).
		\end{aligned}
	\end{equation}
	\begin{proof}
		\begin{equation*}
			\begin{aligned}
				&p(X,Y^0, Y^1|T=0)\\
				=&p(Y^0, Y^1|X,T=0)p(X|T=0)\\
				=&p(Y^0, Y^1|X)p(X|T=0). \quad \text{(Unconfoundedness)}
			\end{aligned}
		\end{equation*}
		\begin{equation*}
			\begin{aligned}
				&p(X,Y^0, Y^1|T=1)\\
				=&p(Y^0, Y^1|X,T=1)p(X|T=1)\\
				=&p(Y^0, Y^1|X)p(X|T=1). \quad \text{(Unconfoundedness)}
			\end{aligned}
		\end{equation*}
	\end{proof}
\end{proposition}
Now we can prove Proposition \ref{prop:kl_equal}.
\begin{proof}
	\begin{equation*}
		\begin{aligned}
			&D_{KL}(P_C||P_T)\\
			=&D_{KL}(P(X,Y^0, Y^1|T=0)||P(X,Y^0, Y^1|T=1))\\
			=&\int_{\mathcal{X}}\int_{\mathcal{Y}^0}\int_{\mathcal{Y}^1}p(x,y^0,y^1|T=0) \log \frac{p(x,y^0,y^1|T=0)}{p(x,y^0,y^1|T=1)}dy^1dy^0dx\\
			=&\int_{\mathcal{X}}\int_{\mathcal{Y}^0}\int_{\mathcal{Y}^1}p(y^0,y^1|x)p(x|T=0) \log \frac{p(y^0,y^1|x)p(x|T=0)}{p(y^0,y^1|x)p(x|T=1)}dy^1dy^0dx \quad \text{(By Proposition \ref{prop:prob_decompose})}\\
			=&\int_{\mathcal{X}}\int_{\mathcal{Y}^0}\int_{\mathcal{Y}^1}p(y^0,y^1|x)p(x|T=0) \log \frac{p(x|T=0)}{p(x|T=1)}dy^1dy^0dx\\
			=&\int_{\mathcal{X}}\left(\int_{\mathcal{Y}^0}\int_{\mathcal{Y}^1}p(y^0,y^1|x)dy^1dy^0\right)p(x|T=0) \log \frac{p(x|T=0)}{p(x|T=1)}dx\\
			=&\int_{\mathcal{X}} p(x|T=0) \log \frac{p(x|T=0)}{p(x|T=1)}dx\\
			=&D_{KL}(P(X|T=0)||P(X|T=1))\\
			=&D_{KL}(P^C_X || P^T_X).
		\end{aligned}
	\end{equation*}
	Similarly, it is easy to show $D_{KL}(P_T||P_C)=D_{KL}(P^T_X || P^C_X)$
\end{proof}

\subsection{Proof of Theorem \ref{thm:dual_form}} \label{app:proof_dual}
\begin{lemma}[Theorem 1 in \cite{hu2013kullback}] \label{lemma:dual_form} Let $f_\theta(X)$ denote the loss function of $X$ and it is bounded almost surely. $\theta \in \Theta$ represents the model parameters of the function $f_\theta(X)$. Let $\mathcal{B}_{\epsilon}(P)$ be the uncertainty ball centered at distribution $P$ with ambiguity radius $\epsilon$. Define $\kappa$ as the mass of the distribution $P$ on its essential supremum (Proposition 2 in \cite{hu2013kullback}). Assume $f_\theta(X)$ is bounded and $\log \kappa + \epsilon<0$, then we have
	\begin{align*}
		\mathcal{V}: = \sup_{Q \in \mathcal{B}_{\epsilon}(P)}\mathbb{E}^{Q}[f_\theta(X)] = \min_{\lambda > 0} \lambda \epsilon + \lambda \log \mathbb{E}^{P}[\exp(f_\theta(X)/\lambda)].
	\end{align*}
\end{lemma}
Our Theorem \ref{thm:dual_form} follows by directly applying the above Lemma \ref{lemma:dual_form}.

\subsection{Proof of Theorem \ref{thm:convergence}} \label{app:proof_convergence}
For notational simplicity, we denote $W=(X,T,Y) \in \mathcal{W}$ and $Z=\hat{\tau}(X)Y$. Assume $Z$ is bounded within the range $\unbar{M}$ and $\bar{M}$. Define the following functions:
\begin{equation*}
	\begin{aligned}
		G_0(\lambda_0;W)=\mathbb{E}[g_0(\lambda_0;W)],& \quad \hat{G}_0(\lambda_0;W)=\frac{1}{n}\sum_{i=1}^{n}g_0(\lambda_0;W_i),\\ &\text{where} \ \ g_0(\lambda_0;W)=(1-T) \exp\left(Z/\lambda_0\right);\\
		G_1(\lambda_1;W)=\mathbb{E}[g_1(\lambda_1;W)],& \quad \hat{G}_1(\lambda_1;W)=\frac{1}{n}\sum_{i=1}^{n}g_1(\lambda_1;W_i),\\ &\text{where} \ \ g_1(\lambda_1;W)=T \exp\left(-Z/\lambda_1\right).
	\end{aligned}
\end{equation*}
Then we have the following lemma that guarantees the convergence for $\hat{G}_0(\lambda_0;W)$ and $\hat{G}_1(\lambda_1;W)$.
\begin{lemma} \label{lemma:convergence of G} Assume $0<\unbar{\lambda} \leq \lambda_0, \lambda_1 \leq \bar{\lambda}$, and $\hat{\tau}(X)Y$ is bounded within the range of $\unbar{M}$ to $\bar{M}$. Then with probability $1-\delta$, we have
	\begin{equation}
		\begin{aligned}
			&\text{If } \  \unbar{M} \leq \bar{M} \leq 0: \\	
			&|\hat{G}_0(\lambda_0;W)-G_0(\lambda_0;W)| \leq \mathcal{O}\left(\sqrt{\frac{2\log\frac{2}{\delta}\left(\exp\left(\bar{M}/\bar{\lambda} \right)\right)^2}{n}}\right);\\
			&|\hat{G}_1(\lambda_1;W)-G_1(\lambda_1;W)| \leq \mathcal{O}\left(\sqrt{\frac{2\log\frac{2}{\delta}\left(\exp\left(-\unbar{M}/\unbar{\lambda} \right)\right)^2}{n}}\right).\\
			&\text{If } \  \unbar{M} \leq 0, \bar{M} \geq 0: \\
			&|\hat{G}_0(\lambda_0;W)-G_0(\lambda_0;W)| \leq \mathcal{O}\left(\sqrt{\frac{2\log\frac{2}{\delta}\left(\exp\left(\bar{M}/\unbar{\lambda} \right)\right)^2}{n}}\right);\\
			&|\hat{G}_1(\lambda_1;W)-G_1(\lambda_1;W)| \leq \mathcal{O}\left(\sqrt{\frac{2\log\frac{2}{\delta}\left(\exp\left(-\unbar{M}/\unbar{\lambda} \right)\right)^2}{n}}\right).\\
			&\text{If } \  0 \leq \unbar{M} \leq \bar{M}: \\
			&|\hat{G}_0(\lambda_0;W)-G_0(\lambda_0;W)| \leq \mathcal{O}\left(\sqrt{\frac{2\log\frac{2}{\delta}\left(\exp\left(\bar{M}/\unbar{\lambda} \right)\right)^2}{n}}\right);\\
			&|\hat{G}_1(\lambda_1;W)-G_1(\lambda_1;W)| \leq \mathcal{O}\left(\sqrt{\frac{2\log\frac{2}{\delta}\left(\exp\left(-\unbar{M}/\bar{\lambda} \right)\right)^2}{n}}\right).
		\end{aligned}
	\end{equation}
	\begin{proof}
		Denote $h_0(W_1,W_2,\dots,W_n)=\frac{1}{n}\sum_{i=1}^{n}g_0(\lambda_0;W_i)$. 
		We notice that $h_0(W_1,W_2,\dots,W_n)$ satisfies the bounded difference inequality:
		\begin{equation*}
			\begin{aligned}
				&\sup_{W_1,\dots,W_n, W_i' \in \mathcal{W}} \left|h_0(W_1,\dots,W_i,\cdots,W_n)-h_0(W_1,\dots,W_i',\cdots,W_n)\right|\\
				&=\sup_{W_i, W_i'  \in \mathcal{W}}\frac{\left|g_0(\lambda_0;W_i)-g_0(\lambda_0;W_i')\right|}{n}\\
				&\leq 2\sup_{W_i  \in \mathcal{W}}\frac{|g_0(\lambda_0;W_i)|}{n} \leq \frac{2\exp\left(\bar{M}/\lambda_0 \right)}{n}.
			\end{aligned}
		\end{equation*}
		Note that $|\hat{G}_0(\lambda_0;W)-G_0(\lambda_0;W)|=|h_0(W_1,W_2,\dots,W_n)-\mathbb{E}[h_0(W_1,W_2,\dots,W_n)]|$. Then using McDiarmid's inequality, for any $\epsilon>0$, we have
		\begin{equation*}
			\begin{aligned}
				&P\left(\left|\hat{G}_0(\lambda_0;W)-G_0(\lambda_0;W)\right|\geq\epsilon\right)\\
				&=P\left(\left|h_0(W_1,W_2,\dots,W_n)-\mathbb{E}[h_0(W_1,W_2,\dots,W_n)]\right|\geq\epsilon\right)\\
				&\leq 2\exp\left(-\frac{2\epsilon^2}{n(\frac{2\exp\left(\bar{M}/\lambda_0\right)}{n})^2}\right)
				=2\exp\left(\frac{-n\epsilon^2}{2\left(\exp\left(\bar{M}/\lambda_0\right)\right)^2}\right).
			\end{aligned}
		\end{equation*}
		For some $\delta>0$, we have
		\begin{equation*}
			\begin{aligned}
				P\left(\left|\hat{G}_0(\lambda_0;W)-G_0(\lambda_0;W)\right|\geq\epsilon\right)
				\leq 2\exp\left(\frac{-n\epsilon^2}{2\left(\exp\left(\bar{M}/\lambda_0\right)\right)^2}\right) \leq \delta.
			\end{aligned}
		\end{equation*}
		This solves $\epsilon$ such that
		\begin{equation*}
			\begin{aligned}
				\epsilon \geq \sqrt{\frac{2\log\frac{2}{\delta}\left(\exp\left(\bar{M}/\lambda_0 \right)\right)^2}{n}}.
			\end{aligned}
		\end{equation*}
		The above inequality should hold for any $\lambda_0$ such that $0<\unbar{\lambda} \leq \lambda_0 \leq \bar{\lambda}$. Therefore, we have
		\begin{equation*}
			\begin{aligned}
				&\text{If} \ \bar{M} \geq 0: \quad	\epsilon \geq \sqrt{\frac{2\log\frac{2}{\delta}\left(\exp\left(\bar{M}/\unbar{\lambda} \right)\right)^2}{n}};\\
				&\text{If} \ \bar{M} \leq 0: \quad	\epsilon \geq \sqrt{\frac{2\log\frac{2}{\delta}\left(\exp\left(\bar{M}/\bar{\lambda} \right)\right)^2}{n}}.\\
			\end{aligned}
		\end{equation*}
		Similarly, denote $h_1(W_1,W_2,\dots,W_n)=\frac{1}{n}\sum_{i=1}^{n}g_1(\lambda_1;W_i)$. We note that $h_1(W_1,W_2,\dots,W_n)$ satisfies the bounded difference inequality:
		\begin{equation*}
			\begin{aligned}
				&\sup_{W_1,\dots,W_n, W_i' \in \mathcal{W}} \left|h_1(W_1,\dots,W_i,\cdots,W_n)-h_1(W_1,\dots,W_i',\cdots,W_n)\right|\\
				&=\sup_{W_i, W_i'  \in \mathcal{W}}\frac{\left|g_1(\lambda_1;W_i)-g_1(\lambda_1;W_i')\right|}{n}\\
				&\leq 2\sup_{W_i  \in \mathcal{W}}\frac{|g_1(\lambda_1;W_i)|}{n} \leq \frac{2\exp\left(-\unbar{M}/\lambda_1 \right)}{n}.
			\end{aligned}
		\end{equation*}
		Then using McDiarmid's inequality, for any $\epsilon>0$, we have
		\begin{equation*}
			\begin{aligned}
				&P\left(\left|\hat{G}_1(\lambda_1;W)-G_1(\lambda_1;W)\right|\geq\epsilon\right)\\
				&=P\left(\left|h_1(W_1,W_2,\dots,W_n)-\mathbb{E}[h_1(W_1,W_2,\dots,W_n)]\right|\geq\epsilon\right)\\
				&\leq 2\exp\left(-\frac{2\epsilon^2}{n(\frac{2\exp\left(-\unbar{M}/\lambda_1\right)}{n})^2}\right)
				=2\exp\left(\frac{-n\epsilon^2}{2\left(\exp\left(-\unbar{M}/\lambda_1\right)\right)^2}\right).
			\end{aligned}
		\end{equation*}
		For some $\delta>0$, we have
		\begin{equation*}
			\begin{aligned}
				P\left(\left|\hat{G}_1(\lambda_1;W)-G_1(\lambda_1;W)\right|\geq\epsilon\right)
				\leq 2\exp\left(\frac{-n\epsilon^2}{2\left(\exp\left(-\unbar{M}/\lambda_1\right)\right)^2}\right) \leq \delta.
			\end{aligned}
		\end{equation*}
		This solves $\epsilon$ such that
		\begin{equation*} \label{eqt:18}
			\begin{aligned}
				\epsilon \geq \sqrt{\frac{2\log\frac{2}{\delta}\left(\exp\left(-\unbar{M}/\lambda_1 \right)\right)^2}{n}}.
			\end{aligned}
		\end{equation*}
		The above inequality should hold for any $\lambda_1$ such that $0<\unbar{\lambda} \leq \lambda_1 \leq \bar{\lambda}$. Therefore, we have
		\begin{equation*} \label{eqt:18}
			\begin{aligned}
				&\text{If} \ \unbar{M} \geq 0: \quad	\epsilon \geq \sqrt{\frac{2\log\frac{2}{\delta}\left(\exp\left(-\unbar{M}/\bar{\lambda} \right)\right)^2}{n}};\\
				&\text{If} \ \unbar{M} \leq 0: \quad	\epsilon \geq \sqrt{\frac{2\log\frac{2}{\delta}\left(\exp\left(-\unbar{M}/\unbar{\lambda} \right)\right)^2}{n}}.\\
			\end{aligned}
		\end{equation*}
	\end{proof}
\end{lemma}
In the following content, we will bound terms $\left|\log(\hat{G}_0(\lambda_0;W))-\log \left(G_0(\lambda_0;W)\right)\right|$ and $\left|\log(\hat{G}_1(\lambda_1;W))-\log \left(G_1(\lambda_1;W)\right)\right|$. Lemma \ref{lemma:log} is useful for bounding these two terms.

\begin{lemma}\label{lemma:log} 
	Let $c$ be a constant. For any $x_1$, $x_2$ such that $x_1, x_2 \geq c >0$, we have 
	\begin{equation}
		\begin{aligned}
			|\log (x_1) - \log (x_2)| \leq \frac{1}{c}|x_1-x_2|
		\end{aligned}
	\end{equation}
	\begin{proof}
		Without loss of generality, assume $0<c \leq x_1 \leq x_2$. We then have
		\begin{equation*}
			\begin{aligned}
				\log (x_2) - \log (x_1) = \log (\frac{x_2}{x_1}) = \log (1+\frac{x_2}{x_1}-1) \leq \frac{x_2}{x_1}-1 = \frac{x_2-x_1}{x_1} \leq \frac{x_2-x_1}{c}.
			\end{aligned}
		\end{equation*}
		Taking the absolute value of both the left-hand side and the right-hand side, we have
		\begin{equation*}
			\begin{aligned}
				|\log (x_1) - \log (x_2)| \leq \frac{1}{c}|x_1-x_2|.
			\end{aligned}
		\end{equation*}
	\end{proof}
\end{lemma}
Next, we introduce Lemma \ref{lemma:bound for G} that bounds terms $\left|\log(\hat{G}_0(\lambda_0;W))-\log \left(G_0(\lambda_0;W)\right)\right|$ and $\left|\log(\hat{G}_1(\lambda_1;W))-\log \left(G_1(\lambda_1;W)\right)\right|$.
\begin{lemma} \label{lemma:bound for G} Let $u$ denote the probability of treat, i.e., $u=P(T=1)$. Assume that $\lambda_0, \lambda_1 \in \Lambda:=[\unbar{\lambda},\bar{\lambda}]$ and $\hat{\tau}(X)Y$ is bounded within $\unbar{M}$ and $\bar{M}$. Then for $n\geq \max\{\frac{2}{u^2}\log \left(\frac{2}{\delta}\right), \frac{2}{(1-u)^2}\log \left(\frac{2}{\delta}\right)\}$, with probability $1-\delta$, we have
	\begin{equation}
		\begin{aligned}
			&\text{If } \  \unbar{M} \leq \bar{M} \leq 0: \\
			&\left|\log(\hat{G}_0(\lambda_0;W))-\log \left(G_0(\lambda_0;W)\right)\right| \leq \frac{2}{\exp (\unbar{M} / \unbar{\lambda}) (1-u)}  \left|\hat{G}_0(\lambda_0;W)-G_0(\lambda_0;W)\right|;\\
			&\left|\log(\hat{G}_1(\lambda_1;W))-\log \left(G_1(\lambda_1;W)\right)\right| \leq \frac{2}{\exp (-\bar{M} / \bar{\lambda}) u}  \left|\hat{G}_1(\lambda_1;W)-G_1(\lambda_1;W)\right|.\\
			&\text{If } \  \unbar{M} \leq 0, \bar{M} \geq 0: \\
			&\left|\log(\hat{G}_0(\lambda_0;W))-\log \left(G_0(\lambda_0;W)\right)\right| \leq \frac{2}{\exp (\unbar{M} / \unbar{\lambda}) (1-u)}  \left|\hat{G}_0(\lambda_0;W)-G_0(\lambda_0;W)\right|;\\
			&\left|\log(\hat{G}_1(\lambda_1;W))-\log \left(G_1(\lambda_1;W)\right)\right|
			\leq \frac{2}{\exp (-\bar{M} / \unbar{\lambda}) u}  
			\left|\hat{G}_1(\lambda_1;W)-G_1(\lambda_1;W)\right|.\\
			&\text{If } \  0 \leq \unbar{M} \leq \bar{M}: \\
			&\left|\log(\hat{G}_0(\lambda_0;W))-\log \left(G_0(\lambda_0;W)\right)\right|
			\leq \frac{2}{\exp (\unbar{M} / \bar{\lambda}) (1-u)}  
			\left|\hat{G}_0(\lambda_0;W)-G_0(\lambda_0;W)\right|;\\
			&\left|\log(\hat{G}_1(\lambda_1;W))-\log \left(G_1(\lambda_1;W)\right)\right|
			\leq \frac{2}{\exp (-\bar{M} / \unbar{\lambda}) u}  
			\left|\hat{G}_1(\lambda_1;W)-G_1(\lambda_1;W)\right|.
		\end{aligned}
	\end{equation}
	\begin{proof}
		First, we bound the term $\left|\log(\hat{G}_0(\lambda_0;W))-\log \left(G_0(\lambda_0;W)\right)\right|$. 
		
		$G_0(\lambda_0;W)$ and $\hat{G}_0(\lambda_0;W)$ are greater than $0$ and bounded because $Z=\hat{\tau}(X)Y$ is bounded within the range $\unbar{M}$ and $\bar{M}$. Therefore, applying Lemma \ref{lemma:log}, we have
		\begin{equation*}
			\begin{aligned}
				&\left|\log(\hat{G}_0(\lambda_0;W))-\log \left(G_0(\lambda_0;W)\right)\right|
				\leq \frac{1}{c}  
				\left|\hat{G}_0(\lambda_0;W)-G_0(\lambda_0;W)\right|,\\
				&\text{where} \; c=\min \left\{ \inf_{\lambda_0 \in \Lambda, W \in \mathcal{W}} \hat{G}_0(\lambda_0;W), \ \inf_{\lambda_0 \in \Lambda, W \in \mathcal{W}} G_0(\lambda_0;W)\right\}.
			\end{aligned}
		\end{equation*}
		Moreover, for any $\lambda_0 \in \Lambda$, we have
		\begin{equation}\label{eqt:20}
			\begin{aligned}
				\text{If } \  \unbar{M} \geq 0:	\quad
				G_0(\lambda_0;W)&=\mathbb{E}[(1-T) \exp (Z / \lambda_0)]=\mathbb{E}[\exp (Z / \lambda_0)|T=0]P(T=0)\\
				&\geq \mathbb{E}[\exp (\unbar{M} / \bar{\lambda})|T=0](1-u)=\exp (\unbar{M} / \bar{\lambda})(1-u); \\ 
				\hat{G}_0(\lambda_0;W)&=\frac{1}{n}\sum_{i=1}^{n}(1-T_i) \exp (Z_i / \lambda_0)\\
				&\geq \frac{1}{n}\sum_{i=1}^{n} (1-T_i) \exp (\unbar{M} / \bar{\lambda}) =\exp (\unbar{M} / \bar{\lambda}) (1-\hat{u}). \\
			\end{aligned}
		\end{equation}
		\begin{equation}\label{eqt:21}
			\begin{aligned}
				\text{If } \  \unbar{M} \leq 0:	\quad
				G_0(\lambda_0;W)&=\mathbb{E}[(1-T) \exp (Z / \lambda_0)]=\mathbb{E}[\exp (Z / \lambda_0)|T=0]P(T=0)\\
				&\geq \mathbb{E}[\exp (\unbar{M} / \unbar{\lambda})|T=0](1-u)=\exp (\unbar{M} / \unbar{\lambda})(1-u); \\ 
				\hat{G}_0(\lambda_0;W)&=\frac{1}{n}\sum_{i=1}^{n}(1-T_i) \exp (Z_i / \lambda_0) \\
				&\geq \frac{1}{n}\sum_{i=1}^{n} (1-T_i) \exp (\unbar{M} / \unbar{\lambda}) =\exp (\unbar{M} / \unbar{\lambda}) (1-\hat{u}).
			\end{aligned}
		\end{equation}
		Given $\hat{u}=\frac{1}{n}\sum_{i=1}^{n}T_i$ and $u=\mathbb{E}[\frac{1}{n}\sum_{i=1}^{n}T_i]$,
		using Hoeffding's inequality, we have
		\begin{equation*}
			\begin{aligned}
				P\left(\left|\frac{1}{n}\sum_{i=1}^{n}(1-T_i)-\mathbb{E}[\frac{1}{n}\sum_{i=1}^{n}(1-T_i)]\right| \geq \frac{\mathbb{E}[\frac{1}{n}\sum_{i=1}^{n}(1-T_i)]}{2}\right)
				\leq 2\exp\left(-\frac{2(\frac{1-u}{2})^2}{n(\frac{1}{n})^2}\right)\leq \delta.
			\end{aligned}
		\end{equation*}
		We can solve $n$ by
		\begin{equation*}
			\begin{aligned}
				2\exp\left(-\frac{n(1-u)^2}{2}\right)\leq \delta \Rightarrow n\geq \frac{2}{(1-u)^2}\log \left(\frac{2}{\delta}\right).
			\end{aligned}
		\end{equation*}
		This indicates that $(1-\hat{u}) \geq (1-u)/2$ with probability $1-\delta$ when $n\geq\frac{2}{(1-u)^2}\log \left(\frac{2}{\delta}\right)$. Combining this with equations \eqref{eqt:20} and \eqref{eqt:21}, with probability $1-\delta$, when $n\geq\frac{2}{(1-u)^2}\log \left(\frac{2}{\delta}\right)$, we have
		\begin{equation*}
			\begin{aligned}
				\text{If } \  \unbar{M} \geq 0: \quad
				&\inf_{\lambda_0 \in \Lambda, W \in \mathcal{W}} G_0(\lambda_0;W) \geq \exp (\unbar{M} / \bar{\lambda})(1-u); \\
				&\inf_{\lambda_0 \in \Lambda, W \in \mathcal{W}} \hat{G}_0(\lambda_0;W) \geq \exp (\unbar{M} / \bar{\lambda}) (1-\hat{u}) \geq \exp (\unbar{M} / \bar{\lambda}) (1-u)/2. \\
			\end{aligned}
		\end{equation*}
		\begin{equation*}
			\begin{aligned}
				\text{If } \  \unbar{M} \leq 0: \quad
				&\inf_{\lambda_0 \in \Lambda, W \in \mathcal{W}} G_0(\lambda_0;W) \geq \exp (\unbar{M} / \unbar{\lambda})(1-u); \\
				&\inf_{\lambda_0 \in \Lambda, W \in \mathcal{W}} \hat{G}_0(\lambda_0;W) \geq \exp (\unbar{M} / \unbar{\lambda}) (1-\hat{u}) \geq \exp (\unbar{M} / \unbar{\lambda}) (1-u)/2.
			\end{aligned}
		\end{equation*}
		Therefore, with probability $1-\delta$, when $n\geq\frac{2}{(1-u)^2}\log \left(\frac{2}{\delta}\right)$, we have
		\begin{equation*}
			\begin{aligned}
				&\text{If } \  \unbar{M} \geq 0: \\
				&\left|\log(\hat{G}_0(\lambda_0;W))-\log \left(G_0(\lambda_0;W)\right)\right|
				\leq \frac{2}{\exp (\unbar{M} / \bar{\lambda}) (1-u)}  
				\left|\hat{G}_0(\lambda_0;W)-G_0(\lambda_0;W)\right|;\\
			\end{aligned}
		\end{equation*}
		\begin{equation*}
			\begin{aligned}
				&\text{If } \  \unbar{M} \leq 0: \\  
				&\left|\log(\hat{G}_0(\lambda_0;W))-\log \left(G_0(\lambda_0;W)\right)\right| \leq \frac{2}{\exp (\unbar{M} / \unbar{\lambda}) (1-u)}  \left|\hat{G}_0(\lambda_0;W)-G_0(\lambda_0;W)\right|.
			\end{aligned}
		\end{equation*}
		Next, we bound the term $\left|\log(\hat{G}_1(\lambda_1;W))-\log \left(G_1(\lambda_1;W)\right)\right|$. $G_1(\lambda_1;W)$ and $\hat{G}_1(\lambda_1;W)$ are greater than $0$ and bounded above. Therefore, applying Lemma \ref{lemma:log}, we have
		\begin{equation*}
			\begin{aligned}
				&\left|\log(\hat{G}_1(\lambda_1;W))-\log \left(G_1(\lambda_1;W)\right)\right|
				\leq \frac{1}{c}  
				\left|\hat{G}_1(\lambda_1;W)-G_1(\lambda_1;W)\right|,\\
				&\text{where} \; c=\min \left\{ \inf_{\lambda_1 \in \Lambda, W \in \mathcal{W}} \hat{G}_1(\lambda_1;W), \ \inf_{\lambda_1 \in \Lambda, W \in \mathcal{W}} G_1(\lambda_1;W)\right\}.
			\end{aligned}
		\end{equation*}
		Moreover, for any $\lambda_1 \in \Lambda$, we have
		\begin{equation}\label{eqt:22}
			\begin{aligned}
				\text{If } \  \bar{M} \geq 0: \quad
				G_1(\lambda_1;W)&=\mathbb{E}[T \exp (-Z / \lambda_1)]=\mathbb{E}[\exp (-Z / \lambda_1)|T=1]P(T=1)\\
				&\geq \mathbb{E}[\exp (-\bar{M} / \unbar{\lambda})|T=1]u=\exp (-\bar{M} / \unbar{\lambda})u; \\ 
				\hat{G}_1(\lambda_1;W)&=\frac{1}{n}\sum_{i=1}^{n}T_i \exp (-Z_i / \lambda_1) \\
				&\geq \frac{1}{n}\sum_{i=1}^{n} T_i \exp (-\bar{M} / \unbar{\lambda}) =\exp (-\bar{M} / \unbar{\lambda}) \hat{u}. \\
			\end{aligned}
		\end{equation}
		\begin{equation}\label{eqt:23}
			\begin{aligned}
				\text{If } \  \bar{M} \leq 0: \quad
				G_1(\lambda_1;W)&=\mathbb{E}[T \exp (-Z / \lambda_1)]=\mathbb{E}[\exp (-Z / \lambda_1)|T=1]P(T=1)\\
				&\geq \mathbb{E}[\exp (-\bar{M} / \bar{\lambda})|T=1]u=\exp (-\bar{M} / \bar{\lambda})u; \\ 
				\hat{G}_1(\lambda_1;W)&=\frac{1}{n}\sum_{i=1}^{n}T_i \exp (-Z_i / \lambda_1) \\
				&\geq \frac{1}{n}\sum_{i=1}^{n} T_i \exp (-\bar{M} / \bar{\lambda}) =\exp (-\bar{M} / \bar{\lambda}) \hat{u}. \\
			\end{aligned}
		\end{equation}
		Given $\hat{u}=\frac{1}{n}\sum_{i=1}^{n}T_i$ and $u=\mathbb{E}[\frac{1}{n}\sum_{i=1}^{n}T_i]$,
		using Hoeffding's inequality, we have
		\begin{equation*}
			\begin{aligned}
				P\left(\left|\frac{1}{n}\sum_{i=1}^{n}T_i-\mathbb{E}[\frac{1}{n}\sum_{i=1}^{n}T_i]\right| \geq \frac{\mathbb{E}[\frac{1}{n}\sum_{i=1}^{n}T_i]}{2}\right)
				\leq 2\exp\left(-\frac{2(\frac{u}{2})^2}{n(\frac{1}{n})^2}\right)\leq \delta.
			\end{aligned}
		\end{equation*}
		We can solve $n$ by
		\begin{equation*}
			\begin{aligned}
				2\exp\left(-\frac{nu^2}{2}\right)\leq \delta \Rightarrow n\geq \frac{2}{u^2}\log \left(\frac{2}{\delta}\right). \label{eqt:compare_u_uhat}
			\end{aligned}
		\end{equation*}
		This indicates that $hat{u} \geq u/2$ with probability $1-\delta$ when $n\geq\frac{2}{u^2}\log \left(\frac{2}{\delta}\right)$. Combining this with equations \eqref{eqt:22} and \eqref{eqt:23}, with probability $1-\delta$, when $n\geq\frac{2}{u^2}\log \left(\frac{2}{\delta}\right)$, we have
		\begin{equation*}
			\begin{aligned}
				\text{If } \  \bar{M} \geq 0: \quad
				&\inf_{\lambda_1 \in \Lambda, W \in \mathcal{W}} G_1(\lambda_1;W) \geq \exp (-\bar{M} / \unbar{\lambda})u; \\
				&\inf_{\lambda_1 \in \Lambda, W \in \mathcal{W}} \hat{G}_1(\lambda_1;W) \geq \exp (-\bar{M} / \unbar{\lambda})\hat{u} \geq \exp (-\bar{M} / \unbar{\lambda}) u/2.
			\end{aligned}
		\end{equation*}
		\begin{equation*}
			\begin{aligned}		
				\text{If } \  \bar{M} \leq 0: \quad
				&\inf_{\lambda_1 \in \Lambda, W \in \mathcal{W}} G_1(\lambda_1;W) \geq \exp (-\bar{M} / \bar{\lambda})u; \\
				&\inf_{\lambda_1 \in \Lambda, W \in \mathcal{W}} \hat{G}_1(\lambda_1;W) \geq \exp (-\bar{M} / \bar{\lambda}) \hat{u} \geq \exp (\bar{M} / \bar{\lambda}) u/2. 
			\end{aligned}
		\end{equation*}
		
		Therefore, with probability $1-\delta$, when $n\geq\frac{2}{u^2}\log \left(\frac{2}{\delta}\right)$, we have
		\begin{equation*}
			\begin{aligned}
				&\text{If } \  \bar{M} \geq 0: \\  
				&\left|\log(\hat{G}_1(\lambda_1;W))-\log \left(G_1(\lambda_1;W)\right)\right|
				\leq \frac{2}{\exp (-\bar{M} / \unbar{\lambda}) u}  
				\left|\hat{G}_1(\lambda_1;W)-G_1(\lambda_1;W)\right|;
			\end{aligned}
		\end{equation*}
		\begin{equation*}
			\begin{aligned}
				&\text{If } \  \bar{M} \leq 0: \\  
				&\left|\log(\hat{G}_1(\lambda_1;W))-\log \left(G_1(\lambda_1;W)\right)\right| \leq \frac{2}{\exp (-\bar{M} / \bar{\lambda}) u}  \left|\hat{G}_1(\lambda_1;W)-G_1(\lambda_1;W)\right|.
			\end{aligned}
		\end{equation*}
		This completes the proof of Lemma \ref{lemma:bound for G}.
	\end{proof}
\end{lemma}

Additionally, the following Lemma \ref{lemma:log_u} provides the bound of $|\log(\hat{u}) - \log(u)|$.
\begin{lemma} \label{lemma:log_u} Let $\hat{u}=\frac{1}{n}\sum_{i=1}^{n}T_i$ and $u=\mathbb{E}[\frac{1}{n}\sum_{i=1}^{n}T_i]$. For $n\geq\frac{2}{u^2}\log \left(\frac{2}{\delta}\right)$, with probability $1-\delta$, we have
	\begin{equation}
		\begin{aligned}
			|\log(\hat{u}) - \log(u)|\leq \mathcal{O}\left(\sqrt{\frac{2\log(\frac{2}{\delta})}{nu^2}}\right).
		\end{aligned}
	\end{equation}
	\begin{proof}
		Using Hoeffding's inequality, we have
		\begin{equation*}
			\begin{aligned}
				&P(|\hat{u}-u|\geq \epsilon)=P\left(\left|\frac{1}{n}\sum_{i=1}^{n}T_i-\mathbb{E}[\frac{1}{n}\sum_{i=1}^{n}T_i]\right| \geq \epsilon \right)
				\leq 2\exp\left(-2n\epsilon^2\right),\\
				&2\exp\left(-2n\epsilon^2\right) \leq \delta \quad \text{solves} \quad \epsilon \geq \sqrt{\frac{\log(\frac{2}{\delta})}{2n}}.
			\end{aligned}
		\end{equation*}
		Notably, using the results in the previous lemma, we know for $n\geq\frac{2}{u^2}\log \left(\frac{2}{\delta}\right)$, $\hat{u}\geq u/2$. Therefore, we have 
		\begin{equation*}
			\begin{aligned}
				|\log(\hat{u}) - \log(u)|&\leq \frac{1}{\min\{\hat{u},u\}}|\hat{u}-u|. \quad \text{(By Lemma \ref{lemma:log})}\\
				&\leq \frac{2}{u}|\hat{u}-u|\leq\frac{2}{u}\mathcal{O}\left(\sqrt{\frac{\log(\frac{2}{\delta})}{2n}}\right)=\mathcal{O}\left(\sqrt{\frac{2\log(\frac{2}{\delta})}{nu^2}}\right).
			\end{aligned}
		\end{equation*}
	\end{proof}
\end{lemma}
In the following, we will bound the term $|\hat{\mathcal{V}}(\hat{\tau})-\mathcal{V}(\hat{\tau})|$ using above lemmas. We first define functions $F_0(\lambda_0)$, $\hat{F}_0(\lambda_0)$, $F_1(\lambda_1)$, and $\hat{F}_1(\lambda_1)$:
\begin{equation*}
	\begin{aligned}
		F_0(\lambda_0)&=\lambda_0 \epsilon_0 + \lambda_0\log (\mathbb{E}^{P_C}[\exp (\hat{\tau}(X)Y / \lambda_0)])\\
		&=\lambda_0 \epsilon_0 + \lambda_0 \log \left(\frac{1}{1-u}\mathbb{E}[(1-T) \exp (\hat{\tau}(X)Y / \lambda_0)] \right);\\
		\hat{F}_0(\lambda_0)&= \lambda_0 \epsilon_0 + \lambda_0 \log (\frac{1}{n_c}\sum_{i=1}^{n}(1-T_i) \exp (\hat{\tau}(X_i)Y_i / \lambda_0))\\
		&=\lambda_0 \epsilon_0 + \lambda_0 \log \left(\frac{1}{n(1-\hat{u})}\sum_{i=1}^{n}(1-T_i) \exp (\hat{\tau}(X_i)Y_i / \lambda_0)\right).\\
		F_1(\lambda_1)&=\lambda_1 \epsilon_1 + \lambda_1\log (\mathbb{E}^{P_T}[\exp (-\hat{\tau}(X)Y / \lambda_1)])\\
		&=\lambda_1 \epsilon_1 + \lambda_1 \log \left(\frac{1}{u}\mathbb{E}[T \exp (-\hat{\tau}(X)Y / \lambda_1)] \right);\\
		\hat{F}_1(\lambda_1)&= \lambda_1 \epsilon_1 + \lambda_1 \log (\frac{1}{n_t}\sum_{i=1}^{n}T_i \exp (-\hat{\tau}(X_i)Y_i / \lambda_1))\\
		&=\lambda_1 \epsilon_1 + \lambda_1 \log \left(\frac{1}{n \hat{u}}\sum_{i=1}^{n}T_i \exp (-\hat{\tau}(X_i)Y_i / \lambda_1)\right).
	\end{aligned}
\end{equation*}

The following Lemma \ref{lemma:bound for F} bounds the term $|\hat{F}(\lambda)-F(\lambda)|$.
\begin{lemma} \label{lemma:bound for F} Let $u:=P(T=1)$. Assuming that $0< \unbar{\lambda} \leq \lambda \leq \bar{\lambda}$ and $\hat{\tau}(X)Y$ is bounded within the range of $\unbar{M}$ to $\bar{M}$. Define $C_{exp}=\mathbf{1}_{\{\unbar{M} \leq \bar{M} \leq 0\}}\exp\left(\bar{M}/\bar{\lambda} - \unbar{M}/\unbar{\lambda} \right)+\mathbf{1}_{\{\unbar{M} \leq 0, \bar{M} \geq 0\}}\exp\left(\bar{M}/\unbar{\lambda} - \unbar{M}/\unbar{\lambda} \right)+\mathbf{1}_{\{0 \leq \unbar{M} \leq \bar{M}\}}\exp\left(\bar{M}/\unbar{\lambda} - \unbar{M}/\bar{\lambda}\right)$. For $n \geq 2/u^2 \log (2/\delta)$, with probability $1-\delta$, we have
	\begin{equation}
		\begin{aligned}
			&|\hat{F}_0(\lambda_0)-F_0(\lambda_0)| \leq \mathcal{O}\left(\sqrt{\frac{8 \lambda_0^2 \log\frac{2}{\delta}}{n(1-u)^2} C_{exp}^2 }\right) + \mathcal{O}\left(\sqrt{\frac{2\lambda_0^2\log(\frac{2}{\delta})}{n(1-u)^2}} \right);\\
			&|\hat{F}_1(\lambda_1)-F_1(\lambda_1)| \leq \mathcal{O}\left(\sqrt{\frac{8 \lambda_1^2 \log\frac{2}{\delta}}{nu^2} C_{exp}^2 }\right) + \mathcal{O}\left(\sqrt{\frac{2\lambda_1^2\log(\frac{2}{\delta})}{nu^2}} \right).
		\end{aligned}
	\end{equation}
	\begin{proof}
		\begin{equation*}
			\begin{aligned}
				&|\hat{F}_0(\lambda_0)-F_0(\lambda_0)|\\
				&=\left|\lambda_0\left(\log \left(\frac{1}{1-u}\mathbb{E}[(1-T) \exp (\hat{\tau}(X)Y / \lambda_0)] \right) -\log \left( \frac{1}{n(1-\hat{u})}\sum_{i=1}^{n}(1-T_i) \exp (\hat{\tau}(X_i)Y_i / \lambda_0) \right) \right)\right|\\
				&=\lambda_0\left|\log \left(\mathbb{E}[(1-T) \exp (\hat{\tau}(X)Y / \lambda_0)] \right) -\log \left( \frac{1}{n}\sum_{i=1}^{n}(1-T_i) \exp (\hat{\tau}(X_i)Y_i / \lambda_0) \right) +\log(1-\hat{u}) - \log(1-u)\right|\\
				&\leq \lambda_0 \left|\log \left(\mathbb{E}[(1-T) \exp (\hat{\tau}(X)Y / \lambda_0)] \right) -\log \left( \frac{1}{n}\sum_{i=1}^{n}(1-T_i) \exp (\hat{\tau}(X_i)Y_i / \lambda_0) \right)\right| + \lambda_0 \left|\log(1-\hat{u}) - \log(1-u) \right|.
			\end{aligned}
		\end{equation*}
		\begin{equation*}
			\begin{aligned}	
				&\text{If } \  \unbar{M} \leq \bar{M} \leq 0: \\			
				&|\hat{F}_0(\lambda_0)-F_0(\lambda_0)| \\
				& \leq \frac{2\lambda_0}{\exp (\unbar{M} / \unbar{\lambda}) (1-u)}  \left|\hat{G}_0(\lambda_0;W)-G_0(\lambda_0;W)\right| + \lambda_0 \left|\log(1-\hat{u}) - \log(1-	u) \right| \quad \text{(By Lemma \ref{lemma:bound for G})}\\
				& \leq \mathcal{O}\left(\sqrt{\frac{8 \lambda_0^2 \log\frac{2}{\delta}}{n(1-u)^2} \left(\exp\left(\bar{M}/\bar{\lambda} - \unbar{M}/\unbar{\lambda} \right)\right)^2 }\right) + \mathcal{O}\left(\sqrt{\frac{2\lambda_0^2\log(\frac{2}{\delta})}{n(1-u)^2}} \right)  \quad \text{(By Lemma \ref{lemma:convergence of G} and Lemma \ref{lemma:log_u})} \\
				&\text{If } \  \unbar{M} \leq 0, \bar{M} \geq 0: \\		
				&|\hat{F}_0(\lambda_0)-F_0(\lambda_0)| \\
				& \leq \frac{2 \lambda_0}{\exp (\unbar{M} / \unbar{\lambda}) (1-u)}  \left|\hat{G}_0(\lambda_0;W)-G_0(\lambda_0;W)\right| + \lambda_0 \left|\log(1-\hat{u}) - \log(1-	u) \right| \quad \text{(By Lemma \ref{lemma:bound for G})}\\
				& \leq \mathcal{O}\left(\sqrt{\frac{8 \lambda_0^2 \log\frac{2}{\delta}}{n(1-u)^2} \left(\exp\left(\bar{M}/\unbar{\lambda} - \unbar{M}/\unbar{\lambda} \right)\right)^2 }\right) + \mathcal{O}\left(\sqrt{\frac{2\lambda_0^2\log(\frac{2}{\delta})}{n(1-u)^2}} \right)  \quad \text{(By Lemma \ref{lemma:convergence of G} and Lemma \ref{lemma:log_u})} \\
				&\text{If } \  0 \leq \unbar{M} \leq \bar{M}: \\
				&|\hat{F}_0(\lambda_0)-F_0(\lambda_0)| \\
				&\leq \frac{2 \lambda_0}{\exp (\unbar{M} / \bar{\lambda}) (1-u)}  \left|\hat{G}_0(\lambda_0;W)-G_0(\lambda_0;W)\right| + \lambda_0 \left|\log(1-\hat{u}) - \log(1-	u) \right| \quad \text{(By Lemma \ref{lemma:bound for G})}\\
				& \leq \mathcal{O}\left(\sqrt{\frac{8 \lambda_0^2 \log\frac{2}{\delta}}{n(1-u)^2} \left(\exp\left(\bar{M}/\unbar{\lambda} - \unbar{M}/\bar{\lambda} \right)\right)^2 }\right) + \mathcal{O}\left(\sqrt{\frac{2\lambda_0^2\log(\frac{2}{\delta})}{n(1-u)^2}} \right)  \quad \text{(By Lemma \ref{lemma:convergence of G} and Lemma \ref{lemma:log_u})} \\
			\end{aligned}
		\end{equation*}
		
		\begin{equation*}
			\begin{aligned}
				&|\hat{F}_1(\lambda_1)-F_1(\lambda_1)|\\
				&=\left|\lambda_1\left(\log \left(\frac{1}{u}\mathbb{E}[T \exp (-\hat{\tau}(X)Y / \lambda_1)] \right) -\log \left( \frac{1}{n\hat{u}}\sum_{i=1}^{n}T_i \exp (\hat{\tau}(X_i)Y_i / \lambda_0) \right) \right)\right|\\
				&=\lambda_1\left|\log \left(\mathbb{E}[T \exp (\hat{\tau}(X)Y / \lambda_1)] \right) -\log \left( \frac{1}{n}\sum_{i=1}^{n}T_i \exp (\hat{\tau}(X_i)Y_i / \lambda_1) \right) +\log(\hat{u}) - \log(u)\right|\\
				&\leq \lambda_1 \left|\log \left(\mathbb{E}[T \exp (\hat{\tau}(X)Y / \lambda_1)] \right) -\log \left( \frac{1}{n}\sum_{i=1}^{n}T_i \exp (\hat{\tau}(X_i)Y_i / \lambda_1) \right)\right| + \lambda_1 \left|\log(\hat{u}) - \log(u) \right|.
			\end{aligned}
		\end{equation*}
		\begin{equation*}
			\begin{aligned}	
				&\text{If } \  \unbar{M} \leq \bar{M} \leq 0: \\			
				&|\hat{F}_1(\lambda_1)-F_1(\lambda_1)| \\
				&\leq \frac{2\lambda_1}{\exp (-\bar{M} / \bar{\lambda}) u}  \left|\hat{G}_0(\lambda_0;W)-G_0(\lambda_0;W)\right| + \lambda_1 \left|\log(\hat{u}) - \log(u) \right| \quad \text{(By Lemma \ref{lemma:bound for G})}\\
				& \leq \mathcal{O}\left(\sqrt{\frac{8 \lambda_1^2 \log\frac{2}{\delta}}{nu^2} \left(\exp\left(\bar{M}/\bar{\lambda} - \unbar{M}/\unbar{\lambda} \right)\right)^2 }\right) + \mathcal{O}\left(\sqrt{\frac{2\lambda_1^2\log(\frac{2}{\delta})}{nu^2}} \right)  \quad \text{(By Lemma \ref{lemma:convergence of G} and Lemma \ref{lemma:log_u})} \\
				&\text{If } \  \unbar{M} \leq 0, \bar{M} \geq 0: \\		
				&|\hat{F}_1(\lambda_1)-F_1(\lambda_1)| \\
				&\leq \frac{2 \lambda_1}{\exp (-\bar{M} / \unbar{\lambda}) u}  \left|\hat{G}_1(\lambda_1;W)-G_1(\lambda_1;W)\right| + \lambda_1 \left|\log(\hat{u}) - \log(u) \right| \quad \text{(By Lemma \ref{lemma:bound for G})}\\
				& \leq \mathcal{O}\left(\sqrt{\frac{8 \lambda_1^2 \log\frac{2}{\delta}}{nu^2} \left(\exp\left(\bar{M}/\unbar{\lambda} - \unbar{M}/\unbar{\lambda} \right)\right)^2 }\right) + \mathcal{O}\left(\sqrt{\frac{2\lambda_1^2\log(\frac{2}{\delta})}{nu^2}} \right)  \quad \text{(By Lemma \ref{lemma:convergence of G} and Lemma \ref{lemma:log_u})} \\
				&\text{If } \  0 \leq \unbar{M} \leq \bar{M}: \\
				&|\hat{F}_1(\lambda_1)-F_1(\lambda_1)| \\
				&\leq \frac{2 \lambda_1}{\exp (-\bar{M} / \unbar{\lambda}) u}  \left|\hat{G}_1(\lambda_1;W)-G_1(\lambda_1;W)\right| + \lambda_1 \left|\log(\hat{u}) - \log(u) \right| \quad \text{(By Lemma \ref{lemma:bound for G})}\\
				& \leq \mathcal{O}\left(\sqrt{\frac{8 \lambda_1^2 \log\frac{2}{\delta}}{nu^2} \left(\exp\left(\bar{M}/\unbar{\lambda} - \unbar{M}/\bar{\lambda} \right)\right)^2 }\right) + \mathcal{O}\left(\sqrt{\frac{2\lambda_1^2\log(\frac{2}{\delta})}{nu^2}} \right)  \quad \text{(By Lemma \ref{lemma:convergence of G} and Lemma \ref{lemma:log_u})} \\
			\end{aligned}
		\end{equation*}
		
	\end{proof}
\end{lemma}
Now, we can prove the result in Theorem \ref{thm:convergence}. 
\begin{proof}
	Let $\hat{\lambda}_0=\arg\min_{\lambda}\hat{F}_0(\lambda_0)$, $\lambda_0^*=\arg\min_{\lambda_0} F_0(\lambda_0)$, $\hat{\lambda}_1=\arg\min_{\lambda}\hat{F}_1(\lambda_1)$ and $\lambda_1^*=\arg\min_{\lambda_1} F_1(\lambda_1)$. Then we have
	\begin{equation*}
		\begin{aligned}
			\mathcal{V}^0(\hat{\tau})-\hat{\mathcal{V}}^0(\hat{\tau})&=F_0(\lambda_0^*)-\hat{F}_0(\hat{\lambda}_0)\\
			&=F_0(\lambda_0^*)-\hat{F}_0(\hat{\lambda_0})+F_0(\hat{\lambda}_0)-F_0(\hat{\lambda}_0)\\
			&=F_0(\hat{\lambda}_0)-\hat{F}_0(\hat{\lambda}_0) + F_0(\lambda_0^*) - F_0(\hat{\lambda}_0)\\
			&\leq |F_0(\hat{\lambda}_0)-\hat{F}_0(\hat{\lambda}_0)| + 0\\
			&\leq \sup_{\lambda_0}|F_0(\lambda_0)-\hat{F}_0(\lambda_0)|.
		\end{aligned}
	\end{equation*}
	\begin{equation*}
		\begin{aligned}
			\hat{\mathcal{V}}^0(\hat{\tau})-\mathcal{V}^0(\hat{\tau})&=\hat{F}_0(\hat{\lambda}_0)-F_0(\lambda_0^*)\\
			&=\hat{F}_0(\hat{\lambda}_0)-F_0(\lambda_0^*)+\hat{F}_0(\lambda_0^*)-\hat{F}_0(\lambda_0^*)\\
			&=\hat{F}_0(\lambda_0^*)-F_0(\lambda_0^*)+\hat{F}_0(\hat{\lambda}_0)-\hat{F}_0(\lambda_0^*)\\
			&\leq |\hat{F}_0(\lambda_0^*)-F_0(\lambda_0^*)| + 0\\
			&\leq \sup_{\lambda_0}|\hat{F}_0(\lambda_0)-F_0(\lambda_0)|.
		\end{aligned}
	\end{equation*}
	\begin{equation*}
		\begin{aligned}
			\mathcal{V}^1(\hat{\tau})-\hat{\mathcal{V}}^1(\hat{\tau})&=F_1(\lambda_1^*)-\hat{F}_1(\hat{\lambda}_1)\\
			&=F_1(\lambda_1^*)-\hat{F}_1(\hat{\lambda_1})+F_1(\hat{\lambda}_1)-F_1(\hat{\lambda}_1)\\
			&=F_1(\hat{\lambda}_1)-\hat{F}_1(\hat{\lambda}_1) + F_1(\lambda_1^*) - F_1(\hat{\lambda}_1)\\
			&\leq |F_1(\hat{\lambda}_1)-\hat{F}_1(\hat{\lambda}_1)| + 0\\
			&\leq \sup_{\lambda_1}|F_1(\lambda_1)-\hat{F}_1(\lambda_1)|.
		\end{aligned}
	\end{equation*}
	\begin{equation*}
		\begin{aligned}
			\hat{\mathcal{V}}^1(\hat{\tau})-\mathcal{V}^1(\hat{\tau})&=\hat{F}_1(\hat{\lambda}_1)-F_1(\lambda_1^*)\\
			&=\hat{F}_1(\hat{\lambda}_1)-F_1(\lambda_1^*)+\hat{F}_1(\lambda_1^*)-\hat{F}_1(\lambda_1^*)\\
			&=\hat{F}_1(\lambda_1^*)-F_1(\lambda_1^*)+\hat{F}_1(\hat{\lambda}_1)-\hat{F}_1(\lambda_1^*)\\
			&\leq |\hat{F}_1(\lambda_1^*)-F_1(\lambda_1^*)| + 0\\
			&\leq \sup_{\lambda_1}|\hat{F}_1(\lambda_1)-F_1(\lambda_1)|.
		\end{aligned}
	\end{equation*}
	Therefore, we have
	\begin{equation*}
		\begin{aligned}	
			&\text{If } \  \unbar{M} \leq \bar{M} \leq 0: \\			
			&|\hat{\mathcal{V}}^0(\hat{\tau})-\mathcal{V}^0(\hat{\tau})|
			\leq \sup_{\lambda}|\hat{F}(\lambda)-F(\lambda)|
			\leq \mathcal{O}\left(\sqrt{\frac{8 \bar{\lambda}^2 \log\frac{2}{\delta}}{n(1-u)^2} \left(\exp\left(\bar{M}/\bar{\lambda} - \unbar{M}/\unbar{\lambda} \right)\right)^2 }\right) + \mathcal{O}\left(\sqrt{\frac{2\bar{\lambda}^2\log(\frac{2}{\delta})}{n(1-u)^2}} \right);\\
			&|\hat{\mathcal{V}}^1(\hat{\tau})-\mathcal{V}^1(\hat{\tau})|
			\leq \sup_{\lambda}|\hat{F}(\lambda)-F(\lambda)|
			\leq \mathcal{O}\left(\sqrt{\frac{8 \bar{\lambda}^2 \log\frac{2}{\delta}}{nu^2} \left(\exp\left(\bar{M}/\bar{\lambda} - \unbar{M}/\unbar{\lambda} \right)\right)^2 }\right) + \mathcal{O}\left(\sqrt{\frac{2\bar{\lambda}^2\log(\frac{2}{\delta})}{nu^2}} \right).\\
			&\text{If } \  \unbar{M} \leq 0, \bar{M} \geq 0: \\		
			&|\hat{\mathcal{V}}^0(\hat{\tau})-\mathcal{V}^0(\hat{\tau})|
			\leq \sup_{\lambda}|\hat{F}(\lambda)-F(\lambda)|
			\leq  \mathcal{O}\left(\sqrt{\frac{8 \bar{\lambda}^2 \log\frac{2}{\delta}}{n(1-u)^2} \left(\exp\left(\bar{M}/\unbar{\lambda} - \unbar{M}/\unbar{\lambda} \right)\right)^2 }\right) + \mathcal{O}\left(\sqrt{\frac{2\bar{\lambda}^2\log(\frac{2}{\delta})}{n(1-u)^2}} \right);\\
			&|\hat{\mathcal{V}}^1(\hat{\tau})-\mathcal{V}^1(\hat{\tau})|
			\leq \sup_{\lambda}|\hat{F}(\lambda)-F(\lambda)|
			\leq  \mathcal{O}\left(\sqrt{\frac{8 \bar{\lambda}^2 \log\frac{2}{\delta}}{nu^2} \left(\exp\left(\bar{M}/\unbar{\lambda} - \unbar{M}/\unbar{\lambda} \right)\right)^2 }\right) + \mathcal{O}\left(\sqrt{\frac{2\bar{\lambda}^2\log(\frac{2}{\delta})}{nu^2}} \right).\\
			&\text{If } \  0 \leq \unbar{M} \leq \bar{M}: \\
			&|\hat{\mathcal{V}}^0(\hat{\tau})-\mathcal{V}^0(\hat{\tau})|
			\leq \sup_{\lambda}|\hat{F}(\lambda)-F(\lambda)|
			\leq  \mathcal{O}\left(\sqrt{\frac{8 \bar{\lambda}^2 \log\frac{2}{\delta}}{n(1-u)^2} \left(\exp\left(\bar{M}/\unbar{\lambda} - \unbar{M}/\bar{\lambda} \right)\right)^2 }\right) + \mathcal{O}\left(\sqrt{\frac{2\bar{\lambda}^2\log(\frac{2}{\delta})}{n(1-u)^2}} \right);\\
			&|\hat{\mathcal{V}}^1(\hat{\tau})-\mathcal{V}^1(\hat{\tau})|
			\leq \sup_{\lambda}|\hat{F}(\lambda)-F(\lambda)|
			\leq  \mathcal{O}\left(\sqrt{\frac{8 \bar{\lambda}^2 \log\frac{2}{\delta}}{nu^2} \left(\exp\left(\bar{M}/\unbar{\lambda} - \unbar{M}/\bar{\lambda} \right)\right)^2 }\right) + \mathcal{O}\left(\sqrt{\frac{2\bar{\lambda}^2\log(\frac{2}{\delta})}{nu^2}} \right).
		\end{aligned}
	\end{equation*}
	Finally, we have
	\begin{equation*}
		\begin{aligned}			
			|\hat{\mathcal{V}}_t(\hat{\tau})-\mathcal{V}_t(\hat{\tau})|
			&\leq \mathcal{O}\left(\sqrt{\frac{8 \bar{\lambda}^2 \log\frac{2}{\delta}}{nu_t^2} C_{exp}^2 }\right) + \mathcal{O}\left(\sqrt{\frac{2\bar{\lambda}^2\log(\frac{2}{\delta})}{nu_t^2}} \right).
		\end{aligned}
	\end{equation*}
	Note that $u_1=P(T=1)$ and $u_0=P(T=0)$. $C_{exp}=\mathbf{1}_{\{\unbar{M} \leq \bar{M} \leq 0\}}\exp\left(\bar{M}/\bar{\lambda} - \unbar{M}/\unbar{\lambda} \right)+\mathbf{1}_{\{\unbar{M} \leq 0, \bar{M} \geq 0\}}\exp\left(\bar{M}/\unbar{\lambda} - \unbar{M}/\unbar{\lambda} \right)+\mathbf{1}_{\{0 \leq \unbar{M} \leq \bar{M}\}}\exp\left(\bar{M}/\unbar{\lambda} - \unbar{M}/\bar{\lambda}\right)$.
	
\end{proof}

\section{Additional Materials}
\subsection{Additional Explanations}\label{app:additional_explanation}
\paragraph{Q1. Why DRM can select CATE estimators that are robust to the uncertainty in PEHE caused by selection bias and unobserved confounders?}
In Section 3.1 and Section 4.1, we have presented theoretical explanations for the reason why DRM can measure a CATE estimator's robustness against selection bias and unobserved confounding. Below we will explain it more specifically.

In causal inference, all the CATE estimators are constructed on the observational factual data. But how reliable the CATE estimator that learned on factual data is? This question can be never known unless we have the knowledge of the oracle PEHE. As shown in equation \eqref{eqt:decompose}, we know that the PEHE is equal to two $\hat{\tau}$-dependent terms, $\mathbb{E}[\hat{\tau}(X)Y^t|T=t]$ and $\mathbb{E}[\hat{\tau}(X)Y^t|T=1-t]$. Unfortunately, $\mathbb{E}[\hat{\tau}(X)Y^t|T=1-t]$ is uncomputable empirically because we can only observe the factual distribution $P^F=P(X,Y^t|T=t)$ but not the counterfactual distribution $P^{CF}=P(X,Y^t|T=1-t)$. The unobserved counterfactual distribution can be regarded as an uncertain distribution varying around the observed and certain factual distribution $P^{F}$. If we could assume a "God's perspective" and observe $P^{CF}$ directly, the counterfactual distribution will be certain - like a quantum world! Such an uncertainty in $P^{CF}$ results in the uncertainty in PEHE. Now we will analyze the source of such uncertainty by analyzing the relationship between the uncertain distribution $P^{CF}$ and the certain distribution $P^{F}$ based on equation \eqref{eqt:PF_PCF}:
\begin{equation*}
	\begin{aligned}
		P(X,Y^t|T=1-t)=P(X,Y^t|T=t) \frac{P(Y^t|T=1-t, X)}{P(Y^t|T=t, X)} \frac{P(X|T=1-t)}{P(X|T=t)}.
	\end{aligned}
\end{equation*}
From above, we find the unobservable distribution $P(X,Y^t|T=1-t)$ is equal to the observable distribution $P(X,Y^t|T=t)$ multiplied with $\frac{P(Y^t|T=1-t, X)}{P(Y^t|T=t, X)} \frac{P(X|T=1-t)}{P(X|T=t)}$. In other words, $\frac{p(y^t|T=1-t, x)}{p(y^t|T=t, x)} \frac{p(x|T=1-t)}{p(x|T=t)}$ controls the discrepancy between $P^F$ and $P^{CF}$. Note that if there is no unmeasured confounders, then we have $p(y^t|T=1-t, x)=p(y^t|T=t, x)$; and if there is no selection bias (covariate shift), then we have $p(x|T=1-t)=p(x|T=t)$. Now we understand the root cause of the discrepancy between $P^F$ and $P^{CF}$ (or between $\mathbb{E}[\tau(X)Y^t|T=1-t]$ and $\mathbb{E}[\tau(X)Y^t|T=t]$) lies at the unobserved confounders and selection bias (covariate shift). In the DRM method, the uncertainty caused by potential unobserved confounders and selection bias in PEHE can be further measured as the distributionally robust values $\hat{\mathcal{V}}^1$ and $\hat{\mathcal{V}}^0$. Then the PEHE w.r.t. the CATE estimator $\hat{\tau}$ will be at most $\mathcal{R}^{DRM}(\hat{\tau})$, as shown in equation \eqref{eqt:DRM}. An estimator $\hat{\tau}$ that attains smallest $\mathcal{R}^{DRM}(\hat{\tau})$ by definition reflects the distributional robustness against potential unobserved confounders and selection bias. 

\paragraph{Q2. How to set $\epsilon^*$ when there are unobserved confounders?} When unobserved confounders are present, Proposition 3.6 can also provide guidance for setting $\epsilon^*$. Taking $\epsilon_1^* = D_{KL}(P_C || P_T)$ as an example, we have 
\begin{equation*}
	\begin{aligned}
		&D_{KL}(P_C||P_T)\\
		&=\int_{\mathcal{X}}\int_{\mathcal{Y}^0}\int_{\mathcal{Y}^1}p(y^0,y^1|x,T=0)p(x|T=0) \log \frac{p(y^0,y^1|x,T=0)p(x|T=0)}{p(y^0,y^1|x,T=1)p(x|T=1)}dy^1dy^0dx\\
		&=\int_{\mathcal{X}}\left(\int_{\mathcal{Y}^0}\int_{\mathcal{Y}^1}p(y^0,y^1|x,T=0)dy^1dy^0\right)p(x|T=0) \log \frac{p(x|T=0)}{p(x|T=1)}dx\\
		&\quad+\int_{\mathcal{X}}\int_{\mathcal{Y}^0}\int_{\mathcal{Y}^1}p(y^0,y^1|x,T=0)p(x|T=0) \log \frac{p(y^0,y^1|x,T=0)}{p(y^0,y^1|x,T=1)}dy^1dy^0dx\\
		&=D_{KL}(P(X|T=0)||P(X|T=1))\\
		&\quad +\int_{\mathcal{X}}\int_{\mathcal{Y}^0}\int_{\mathcal{Y}^1}p(y^0,y^1|x,T=0)p(x|T=0) \log \frac{p(y^0,y^1|x,T=0)}{p(y^0,y^1|x,T=1)}dy^1dy^0dx\\
		&>D_{KL}(P^C_X || P^T_X)
	\end{aligned}
\end{equation*}
Therefore, when unobserved confounders present, we can set $\epsilon^*$ to a larger value than the one guided by Proposition \ref{prop:kl_equal}. Simultaneously, as the empirical approximation of $\epsilon_1^* = D_{KL}(P^C_X || P^T_X)$ and $\epsilon_0^* = D_{KL}(P^T_X || P^C_X)$ can be biased, we also suggest set $\epsilon^*$ to a large value than the empirically-computed ones to ensure the ambiguity set is large enough to contain the target distribution. Therefore, we generally set $\epsilon_1^*=D_{KL}(P^C_X || P^T_X)+5.2$ and $\epsilon_0^*=D_{KL}(P^T_X || P^C_X)+5.2$ for all settings in our experiment. Theoretically, a larger $\epsilon^*$ should guarantee the DRM-selected estimator to be more robust, as it allows for a broader range of possible counterfactual distributions in the ambiguity set. However, setting $\epsilon^*$ too large can result in overly conservative estimator selection (similar to the well-known accuracy-robustness tradeoff). Therefore, how to determine a proper ambiguity radius still remains an open challenge in both our work and distributionally robust optimization literature.
\subsection{Hyperparameters}\label{app:hyper}
\begin{itemize}
	\item For linear model, we use LogisticRegressionCV and RidgeCV (both are with 3-fold cross-validation) from sklearn package to tune hyperparameters: Logistic regression: $Cs \in $ \{0.01, 0.1, 1, 10\}; Ridge Regression: $\alpha \in $ \{0.01, 0.1, 1, 10, 100\}.
	\item For Neural Net, we set the hidden layers as [200, 200, 200, 100, 100], each with the ReLU activation function. The model is trained using the Adam optimizer with a learning rate of 0.001, a batch size of 64, and 300 epochs.
	\item For RF, XGBoost, and SVM model, we use AutoML \citep{wang2021flaml, mahajan2022empirical} (with 3-fold cross-validation) from flaml package to tune hyperparameters.
\end{itemize}
\subsection{The Complementary Results}\label{app:sec_all}
First, we would like to emphasize that the experimental results in this final version of the paper differ from those in the original version. We made several revisions based on the feedback from anonymous reviewers: 1) We add Neural Net model to the base ML models and add the U-learner to the meta-learners, increasing the number of CATE estimators from 24 to 36; 2) We adopted AutoML for hyperparameter tuning when training SVM, RF, and XGBoost. All code is available at \url{https://github.com/yiyhuang3/CATE_estimator_selection}.

Below, we prsent the complementary PEHE results for 36 candidate CATE estimators, where the candidate pool contains 4 ML models (LR, SVM, RF, and Neural Net) $\times$ 9 learners (U-, S-, T-, PS-, IPW-, X-, DR-, R-, RA-).
\begin{table}[h]
	\centering
	\caption{Comparison of PEHE for different selectors across Settings A, B, and C (Note that B ($\xi=1$) matches A ($\rho=0.1$)),  with base model for CATE estimator being \{LR, SVM, RF, Net\}. Reported values (mean $\pm$ standard deviation) are computed over 100 experiments. Smaller is better.}
	\resizebox{1\columnwidth}{!}{
		\begin{tabular}{ccccccccc}
			\toprule
			& A ($\rho=0$) & A ($\rho=0.1$) & A ($\rho=0.3$) & B ($\xi=0$) & B ($\xi=2$) & C ($m=0.1$) & C ($m=0.5$) & C ($m=0.9$) \\
			\midrule
			Plug-U & 49.59\small{$\pm$95.07} & 41.93\small{$\pm$61.07} & 36.16\small{$\pm$61.77} & 2.28\small{$\pm$2.32} & 155.24\small{$\pm$291.78} & 42.45\small{$\pm$52.65} & 59.51\small{$\pm$210.54} & 24.37\small{$\pm$26.51} \\
			Plug-S & 5.10\small{$\pm$8.29} & 5.36\small{$\pm$5.84} & 6.29\small{$\pm$5.76} & \textbf{1.99}\small{$\pm$1.41} & 9.18\small{$\pm$11.57} & 5.76\small{$\pm$5.46} & 8.78\small{$\pm$7.55} & 13.45\small{$\pm$9.53} \\
			Plug-PS & 4.80\small{$\pm$7.74} & 5.36\small{$\pm$5.85} & 6.28\small{$\pm$5.75} & \textbf{1.99}\small{$\pm$1.41} & \textbf{9.17}\small{$\pm$11.58} & 5.76\small{$\pm$5.46} & 8.58\small{$\pm$7.40} & 13.45\small{$\pm$9.53} \\
			Plug-T & 60.84\small{$\pm$22.03} & 59.09\small{$\pm$22.88} & 59.39\small{$\pm$21.34} & 12.25\small{$\pm$10.80} & 68.22\small{$\pm$18.14} & 62.90\small{$\pm$19.13} & 48.32\small{$\pm$23.60} & 45.07\small{$\pm$20.63} \\
			Plug-X & 9.82\small{$\pm$10.67} & 10.39\small{$\pm$12.20} & 9.81\small{$\pm$11.30} & 6.52\small{$\pm$10.90} & 14.82\small{$\pm$14.23} & 10.82\small{$\pm$15.24} & 15.80\small{$\pm$15.03} & 20.59\small{$\pm$13.03} \\
			Plug-IPW & 35.09\small{$\pm$28.69} & 38.50\small{$\pm$27.78} & 39.29\small{$\pm$27.48} & 6.19\small{$\pm$7.39} & 61.90\small{$\pm$24.17} & 41.47\small{$\pm$31.37} & 30.02\small{$\pm$22.57} & 29.33\small{$\pm$20.45} \\
			Plug-DR & 44.83\small{$\pm$26.77} & 46.47\small{$\pm$27.02} & 48.23\small{$\pm$26.56} & 5.98\small{$\pm$8.09} & 67.87\small{$\pm$18.94} & 49.61\small{$\pm$33.34} & 33.69\small{$\pm$23.05} & 32.39\small{$\pm$19.28} \\
			Plug-R & 3.64\small{$\pm$5.01} & 5.33\small{$\pm$15.72} & 5.51\small{$\pm$3.75} & 2.19\small{$\pm$2.31} & 13.04\small{$\pm$31.51} & 4.95\small{$\pm$5.38} & 7.56\small{$\pm$7.88} & 10.91\small{$\pm$7.92} \\
			Plug-RA & 58.32\small{$\pm$24.02} & 60.40\small{$\pm$20.13} & 58.63\small{$\pm$22.86} & 8.37\small{$\pm$9.28} & 67.77\small{$\pm$17.82} & 58.91\small{$\pm$19.66} & 45.52\small{$\pm$24.80} & 42.13\small{$\pm$20.35} \\
			Pseudo-DR & 63.07\small{$\pm$22.54} & 63.80\small{$\pm$20.22} & 63.10\small{$\pm$19.41} & 16.51\small{$\pm$23.05} & 73.29\small{$\pm$17.48} & 65.12\small{$\pm$20.14} & 53.87\small{$\pm$26.16} & 53.79\small{$\pm$24.91} \\
			Pseudo-R & 11.57\small{$\pm$27.25} & 16.83\small{$\pm$45.81} & 9.97\small{$\pm$21.23} & 6.49\small{$\pm$20.46} & 18.13\small{$\pm$30.61} & 13.62\small{$\pm$24.78} & 20.96\small{$\pm$30.42} & 30.05\small{$\pm$32.02} \\
			Pseudo-IF & 66.26\small{$\pm$15.20} & 65.21\small{$\pm$16.35} & 66.72\small{$\pm$15.84} & 28.49\small{$\pm$23.55} & 69.01\small{$\pm$16.57} & 63.09\small{$\pm$20.62} & 60.00\small{$\pm$19.18} & 47.40\small{$\pm$20.16} \\
			Random & 7216\small{$\pm$22745} & 6514\small{$\pm$21650} & 4200\small{$\pm$17048} & 1136\small{$\pm$5595} & 7552\small{$\pm$22498} & 3771\small{$\pm$16625} & 6219\small{$\pm$19942} & 3453\small{$\pm$14590} \\
			Fact  & 52.81\small{$\pm$18.01} & 53.58\small{$\pm$19.42} & 55.05\small{$\pm$21.10} & 16.09\small{$\pm$16.50} & 68.50\small{$\pm$27.69} & 51.96\small{$\pm$17.45} & 52.44\small{$\pm$22.51} & 49.16\small{$\pm$24.47} \\
			Matching & 62.57\small{$\pm$21.57} & 64.90\small{$\pm$17.85} & 63.94\small{$\pm$18.72} & 15.10\small{$\pm$22.93} & 72.25\small{$\pm$17.46} & 64.56\small{$\pm$19.59} & 57.87\small{$\pm$24.40} & 48.81\small{$\pm$25.23} \\
			DRM   & \textbf{2.68}\small{$\pm$4.73} & \textbf{3.55}\small{$\pm$5.65} & \textbf{5.28}\small{$\pm$6.37} & 2.14\small{$\pm$1.70} & 18.77\small{$\pm$112.78} & \textbf{4.60}\small{$\pm$9.58} & \textbf{6.44}\small{$\pm$9.73} & \textbf{10.05}\small{$\pm$7.19} \\
			\bottomrule
		\end{tabular}%
	}
	\label{tab:pehe_all}%
\end{table}%

\clearpage


\newpage
\section*{NeurIPS Paper Checklist}

\begin{enumerate}

\item {\bf Claims}
    \item[] Question: Do the main claims made in the abstract and introduction accurately reflect the paper's contributions and scope?
    \item[] Answer: \answerYes{} 
    \item[] Justification: \answerYes{}
    \item[] Guidelines:
    \begin{itemize}
        \item The answer NA means that the abstract and introduction do not include the claims made in the paper.
        \item The abstract and/or introduction should clearly state the claims made, including the contributions made in the paper and important assumptions and limitations. A No or NA answer to this question will not be perceived well by the reviewers. 
        \item The claims made should match theoretical and experimental results, and reflect how much the results can be expected to generalize to other settings. 
        \item It is fine to include aspirational goals as motivation as long as it is clear that these goals are not attained by the paper. 
    \end{itemize}

\item {\bf Limitations}
    \item[] Question: Does the paper discuss the limitations of the work performed by the authors?
    \item[] Answer: \answerYes{} 
    \item[] Justification: \answerYes{}
    \item[] Guidelines:
    \begin{itemize}
        \item The answer NA means that the paper has no limitation while the answer No means that the paper has limitations, but those are not discussed in the paper. 
        \item The authors are encouraged to create a separate "Limitations" section in their paper.
        \item The paper should point out any strong assumptions and how robust the results are to violations of these assumptions (e.g., independence assumptions, noiseless settings, model well-specification, asymptotic approximations only holding locally). The authors should reflect on how these assumptions might be violated in practice and what the implications would be.
        \item The authors should reflect on the scope of the claims made, e.g., if the approach was only tested on a few datasets or with a few runs. In general, empirical results often depend on implicit assumptions, which should be articulated.
        \item The authors should reflect on the factors that influence the performance of the approach. For example, a facial recognition algorithm may perform poorly when image resolution is low or images are taken in low lighting. Or a speech-to-text system might not be used reliably to provide closed captions for online lectures because it fails to handle technical jargon.
        \item The authors should discuss the computational efficiency of the proposed algorithms and how they scale with dataset size.
        \item If applicable, the authors should discuss possible limitations of their approach to address problems of privacy and fairness.
        \item While the authors might fear that complete honesty about limitations might be used by reviewers as grounds for rejection, a worse outcome might be that reviewers discover limitations that aren't acknowledged in the paper. The authors should use their best judgment and recognize that individual actions in favor of transparency play an important role in developing norms that preserve the integrity of the community. Reviewers will be specifically instructed to not penalize honesty concerning limitations.
    \end{itemize}

\item {\bf Theory Assumptions and Proofs}
    \item[] Question: For each theoretical result, does the paper provide the full set of assumptions and a complete (and correct) proof?
    \item[] Answer: \answerYes{}
    \item[] Justification: \answerYes{}
    \item[] Guidelines:
    \begin{itemize}
        \item The answer NA means that the paper does not include theoretical results. 
        \item All the theorems, formulas, and proofs in the paper should be numbered and cross-referenced.
        \item All assumptions should be clearly stated or referenced in the statement of any theorems.
        \item The proofs can either appear in the main paper or the supplemental material, but if they appear in the supplemental material, the authors are encouraged to provide a short proof sketch to provide intuition. 
        \item Inversely, any informal proof provided in the core of the paper should be complemented by formal proofs provided in appendix or supplemental material.
        \item Theorems and Lemmas that the proof relies upon should be properly referenced. 
    \end{itemize}

    \item {\bf Experimental Result Reproducibility}
    \item[] Question: Does the paper fully disclose all the information needed to reproduce the main experimental results of the paper to the extent that it affects the main claims and/or conclusions of the paper (regardless of whether the code and data are provided or not)?
    \item[] Answer: \answerYes{} 
    \item[] Justification: \answerYes{}
    \item[] Guidelines:
    \begin{itemize}
        \item The answer NA means that the paper does not include experiments.
        \item If the paper includes experiments, a No answer to this question will not be perceived well by the reviewers: Making the paper reproducible is important, regardless of whether the code and data are provided or not.
        \item If the contribution is a dataset and/or model, the authors should describe the steps taken to make their results reproducible or verifiable. 
        \item Depending on the contribution, reproducibility can be accomplished in various ways. For example, if the contribution is a novel architecture, describing the architecture fully might suffice, or if the contribution is a specific model and empirical evaluation, it may be necessary to either make it possible for others to replicate the model with the same dataset, or provide access to the model. In general. releasing code and data is often one good way to accomplish this, but reproducibility can also be provided via detailed instructions for how to replicate the results, access to a hosted model (e.g., in the case of a large language model), releasing of a model checkpoint, or other means that are appropriate to the research performed.
        \item While NeurIPS does not require releasing code, the conference does require all submissions to provide some reasonable avenue for reproducibility, which may depend on the nature of the contribution. For example
        \begin{enumerate}
            \item If the contribution is primarily a new algorithm, the paper should make it clear how to reproduce that algorithm.
            \item If the contribution is primarily a new model architecture, the paper should describe the architecture clearly and fully.
            \item If the contribution is a new model (e.g., a large language model), then there should either be a way to access this model for reproducing the results or a way to reproduce the model (e.g., with an open-source dataset or instructions for how to construct the dataset).
            \item We recognize that reproducibility may be tricky in some cases, in which case authors are welcome to describe the particular way they provide for reproducibility. In the case of closed-source models, it may be that access to the model is limited in some way (e.g., to registered users), but it should be possible for other researchers to have some path to reproducing or verifying the results.
        \end{enumerate}
    \end{itemize}

\item {\bf Open access to data and code}
    \item[] Question: Does the paper provide open access to the data and code, with sufficient instructions to faithfully reproduce the main experimental results, as described in supplemental material?
    \item[] Answer: \answerYes{} 
    \item[] Justification: \answerYes{}
    \item[] Guidelines:
    \begin{itemize}
        \item The answer NA means that paper does not include experiments requiring code.
        \item Please see the NeurIPS code and data submission guidelines (\url{https://nips.cc/public/guides/CodeSubmissionPolicy}) for more details.
        \item While we encourage the release of code and data, we understand that this might not be possible, so “No” is an acceptable answer. Papers cannot be rejected simply for not including code, unless this is central to the contribution (e.g., for a new open-source benchmark).
        \item The instructions should contain the exact command and environment needed to run to reproduce the results. See the NeurIPS code and data submission guidelines (\url{https://nips.cc/public/guides/CodeSubmissionPolicy}) for more details.
        \item The authors should provide instructions on data access and preparation, including how to access the raw data, preprocessed data, intermediate data, and generated data, etc.
        \item The authors should provide scripts to reproduce all experimental results for the new proposed method and baselines. If only a subset of experiments are reproducible, they should state which ones are omitted from the script and why.
        \item At submission time, to preserve anonymity, the authors should release anonymized versions (if applicable).
        \item Providing as much information as possible in supplemental material (appended to the paper) is recommended, but including URLs to data and code is permitted.
    \end{itemize}

\item {\bf Experimental Setting/Details}
    \item[] Question: Does the paper specify all the training and test details (e.g., data splits, hyperparameters, how they were chosen, type of optimizer, etc.) necessary to understand the results?
    \item[] Answer: \answerYes{} 
    \item[] Justification: \answerYes{}
    \item[] Guidelines:
    \begin{itemize}
        \item The answer NA means that the paper does not include experiments.
        \item The experimental setting should be presented in the core of the paper to a level of detail that is necessary to appreciate the results and make sense of them.
        \item The full details can be provided either with the code, in appendix, or as supplemental material.
    \end{itemize}

\item {\bf Experiment Statistical Significance}
    \item[] Question: Does the paper report error bars suitably and correctly defined or other appropriate information about the statistical significance of the experiments?
    \item[] Answer: \answerYes{} 
    \item[] Justification: \answerYes{}
    \item[] Guidelines:
    \begin{itemize}
        \item The answer NA means that the paper does not include experiments.
        \item The authors should answer "Yes" if the results are accompanied by error bars, confidence intervals, or statistical significance tests, at least for the experiments that support the main claims of the paper.
        \item The factors of variability that the error bars are capturing should be clearly stated (for example, train/test split, initialization, random drawing of some parameter, or overall run with given experimental conditions).
        \item The method for calculating the error bars should be explained (closed form formula, call to a library function, bootstrap, etc.)
        \item The assumptions made should be given (e.g., Normally distributed errors).
        \item It should be clear whether the error bar is the standard deviation or the standard error of the mean.
        \item It is OK to report 1-sigma error bars, but one should state it. The authors should preferably report a 2-sigma error bar than state that they have a 96\% CI, if the hypothesis of Normality of errors is not verified.
        \item For asymmetric distributions, the authors should be careful not to show in tables or figures symmetric error bars that would yield results that are out of range (e.g. negative error rates).
        \item If error bars are reported in tables or plots, The authors should explain in the text how they were calculated and reference the corresponding figures or tables in the text.
    \end{itemize}

\item {\bf Experiments Compute Resources}
    \item[] Question: For each experiment, does the paper provide sufficient information on the computer resources (type of compute workers, memory, time of execution) needed to reproduce the experiments?
    \item[] Answer: \answerYes{} 
    \item[] Justification: \answerYes{}
    \item[] Guidelines:
    \begin{itemize}
        \item The answer NA means that the paper does not include experiments.
        \item The paper should indicate the type of compute workers CPU or GPU, internal cluster, or cloud provider, including relevant memory and storage.
        \item The paper should provide the amount of compute required for each of the individual experimental runs as well as estimate the total compute. 
        \item The paper should disclose whether the full research project required more compute than the experiments reported in the paper (e.g., preliminary or failed experiments that didn't make it into the paper). 
    \end{itemize}
    
\item {\bf Code Of Ethics}
    \item[] Question: Does the research conducted in the paper conform, in every respect, with the NeurIPS Code of Ethics \url{https://neurips.cc/public/EthicsGuidelines}?
    \item[] Answer: \answerYes{} 
    \item[] Justification: \answerYes{}
    \item[] Guidelines:
    \begin{itemize}
        \item The answer NA means that the authors have not reviewed the NeurIPS Code of Ethics.
        \item If the authors answer No, they should explain the special circumstances that require a deviation from the Code of Ethics.
        \item The authors should make sure to preserve anonymity (e.g., if there is a special consideration due to laws or regulations in their jurisdiction).
    \end{itemize}

\item {\bf Broader Impacts}
    \item[] Question: Does the paper discuss both potential positive societal impacts and negative societal impacts of the work performed?
    \item[] Answer: \answerNA{} 
    \item[] Justification: \answerNA{}
    \item[] Guidelines:
    \begin{itemize}
        \item The answer NA means that there is no societal impact of the work performed.
        \item If the authors answer NA or No, they should explain why their work has no societal impact or why the paper does not address societal impact.
        \item Examples of negative societal impacts include potential malicious or unintended uses (e.g., disinformation, generating fake profiles, surveillance), fairness considerations (e.g., deployment of technologies that could make decisions that unfairly impact specific groups), privacy considerations, and security considerations.
        \item The conference expects that many papers will be foundational research and not tied to particular applications, let alone deployments. However, if there is a direct path to any negative applications, the authors should point it out. For example, it is legitimate to point out that an improvement in the quality of generative models could be used to generate deepfakes for disinformation. On the other hand, it is not needed to point out that a generic algorithm for optimizing neural networks could enable people to train models that generate Deepfakes faster.
        \item The authors should consider possible harms that could arise when the technology is being used as intended and functioning correctly, harms that could arise when the technology is being used as intended but gives incorrect results, and harms following from (intentional or unintentional) misuse of the technology.
        \item If there are negative societal impacts, the authors could also discuss possible mitigation strategies (e.g., gated release of models, providing defenses in addition to attacks, mechanisms for monitoring misuse, mechanisms to monitor how a system learns from feedback over time, improving the efficiency and accessibility of ML).
    \end{itemize}
    
\item {\bf Safeguards}
    \item[] Question: Does the paper describe safeguards that have been put in place for responsible release of data or models that have a high risk for misuse (e.g., pretrained language models, image generators, or scraped datasets)?
    \item[] Answer: \answerNA{} 
    \item[] Justification: \answerNA{}
    \item[] Guidelines:
    \begin{itemize}
        \item The answer NA means that the paper poses no such risks.
        \item Released models that have a high risk for misuse or dual-use should be released with necessary safeguards to allow for controlled use of the model, for example by requiring that users adhere to usage guidelines or restrictions to access the model or implementing safety filters. 
        \item Datasets that have been scraped from the Internet could pose safety risks. The authors should describe how they avoided releasing unsafe images.
        \item We recognize that providing effective safeguards is challenging, and many papers do not require this, but we encourage authors to take this into account and make a best faith effort.
    \end{itemize}

\item {\bf Licenses for existing assets}
    \item[] Question: Are the creators or original owners of assets (e.g., code, data, models), used in the paper, properly credited and are the license and terms of use explicitly mentioned and properly respected?
    \item[] Answer: \answerNA{} 
    \item[] Justification: \answerNA{}
    \item[] Guidelines:
    \begin{itemize}
        \item The answer NA means that the paper does not use existing assets.
        \item The authors should cite the original paper that produced the code package or dataset.
        \item The authors should state which version of the asset is used and, if possible, include a URL.
        \item The name of the license (e.g., CC-BY 4.0) should be included for each asset.
        \item For scraped data from a particular source (e.g., website), the copyright and terms of service of that source should be provided.
        \item If assets are released, the license, copyright information, and terms of use in the package should be provided. For popular datasets, \url{paperswithcode.com/datasets} has curated licenses for some datasets. Their licensing guide can help determine the license of a dataset.
        \item For existing datasets that are re-packaged, both the original license and the license of the derived asset (if it has changed) should be provided.
        \item If this information is not available online, the authors are encouraged to reach out to the asset's creators.
    \end{itemize}

\item {\bf New Assets}
    \item[] Question: Are new assets introduced in the paper well documented and is the documentation provided alongside the assets?
    \item[] Answer: \answerNA{} 
    \item[] Justification: \answerNA{}
    \item[] Guidelines:
    \begin{itemize}
        \item The answer NA means that the paper does not release new assets.
        \item Researchers should communicate the details of the dataset/code/model as part of their submissions via structured templates. This includes details about training, license, limitations, etc. 
        \item The paper should discuss whether and how consent was obtained from people whose asset is used.
        \item At submission time, remember to anonymize your assets (if applicable). You can either create an anonymized URL or include an anonymized zip file.
    \end{itemize}

\item {\bf Crowdsourcing and Research with Human Subjects}
    \item[] Question: For crowdsourcing experiments and research with human subjects, does the paper include the full text of instructions given to participants and screenshots, if applicable, as well as details about compensation (if any)? 
    \item[] Answer: \answerNA{} 
    \item[] Justification: \answerNA{}
    \item[] Guidelines:
    \begin{itemize}
        \item The answer NA means that the paper does not involve crowdsourcing nor research with human subjects.
        \item Including this information in the supplemental material is fine, but if the main contribution of the paper involves human subjects, then as much detail as possible should be included in the main paper. 
        \item According to the NeurIPS Code of Ethics, workers involved in data collection, curation, or other labor should be paid at least the minimum wage in the country of the data collector. 
    \end{itemize}

\item {\bf Institutional Review Board (IRB) Approvals or Equivalent for Research with Human Subjects}
    \item[] Question: Does the paper describe potential risks incurred by study participants, whether such risks were disclosed to the subjects, and whether Institutional Review Board (IRB) approvals (or an equivalent approval/review based on the requirements of your country or institution) were obtained?
    \item[] Answer: \answerNA{} 
    \item[] Justification: \answerNA{}
    \item[] Guidelines:
    \begin{itemize}
        \item The answer NA means that the paper does not involve crowdsourcing nor research with human subjects.
        \item Depending on the country in which research is conducted, IRB approval (or equivalent) may be required for any human subjects research. If you obtained IRB approval, you should clearly state this in the paper. 
        \item We recognize that the procedures for this may vary significantly between institutions and locations, and we expect authors to adhere to the NeurIPS Code of Ethics and the guidelines for their institution. 
        \item For initial submissions, do not include any information that would break anonymity (if applicable), such as the institution conducting the review.
    \end{itemize}

\end{enumerate}

\end{document}